\documentclass[runningheads]{llncs}

\usepackage{eccv}

\usepackage{eccvabbrv}

\usepackage{graphicx}
\usepackage{booktabs}
\usepackage{multirow}

\usepackage[accsupp]{axessibility}  

\usepackage{hyperref}

\usepackage{orcidlink}

\usepackage{array}

\DeclareMathOperator*{\argmin}{arg\,min}

\newcolumntype{M}[1]{>{\centering\arraybackslash}m{#1}}

\begin{document}

\title{Scalable Indoor Novel-View Synthesis using Drone-Captured 360 Imagery with 3D Gaussian Splatting} 

\titlerunning{Scalable Indoor Novel-View Synthesis using Drone-Captured 360 Imagery}

\author{Yuanbo Chen*
\inst{1} \orcidlink{0009-0001-1904-8069} 
\and
Chengyu Zhang*
\inst{1} \orcidlink{0000-0002-7654-6937} 
\and
Jason Wang \and Xuefan Gao \and Avideh Zakhor
\inst{1} \orcidlink{0000-0003-4770-6353}
}

\authorrunning{Y.~Chen et al.}

\institute{UC Berkeley, Berkeley CA 94720, USA \\
\email{\{yuanbo\_chen, chengyu\_zhang, jasoncwang, xuefangao, avz\}@berkeley.edu}}

\maketitle
\def\thefootnote{*}\footnotetext{Both authors contributed equally to the paper.}\def\thefootnote{\arabic{footnote}}

\begin{abstract}
Scene reconstruction and novel-view synthesis for large, complex, multi-story, indoor scenes is a challenging and time-consuming task. Prior methods have utilized drones for data capture and radiance fields for scene reconstruction, both of which present certain challenges. First, in order to capture diverse viewpoints with the drone's front-facing camera, some approaches fly the drone in an unstable zig-zag fashion, which hinders drone-piloting and generates motion blur in the captured data. Secondly, most radiance field methods do not easily scale  to arbitrarily large number of images. 
This paper proposes an efficient and scalable pipeline for indoor novel-view synthesis from drone-captured 360\textdegree \,videos using 3D Gaussian Splatting. 360\textdegree \,cameras capture a wide set of viewpoints, allowing for comprehensive scene capture under a simple straightforward drone trajectory.
To scale our method to large scenes, we devise a divide-and-conquer strategy to automatically split the scene into smaller blocks that can be reconstructed individually and in parallel. We also propose a coarse-to-fine alignment strategy to seamlessly match these blocks together to compose the entire scene. Our experiments demonstrate marked improvement in both reconstruction quality, \ie PSNR and SSIM, and computation time compared to prior approaches.
\end{abstract}

\section{Introduction}
\label{sec:intro}

Novel-view synthesis for indoor environments is a valuable task with wide applications in areas such as heritage preservation~\cite{heritage_2023} and digital twin modeling~\cite{skydio}. Other use cases include generating visual assets for further downstream tasks such as in robot navigation~\cite{byravan2022nerf2real}, AR/VR~\cite{deng2022fovnerf, kang20216, yang2022retargeting}, and medical imaging~\cite{Chen_2023_ICCV, song_deep_2019}. Classical pipelines for novel-view synthesis are as follows: (1) data acquisition, (2) 3D scene reconstruction, and (3) novel-view rendering. Due to the complexities involved in this multi-step process, there are many areas of improvement for better performance, especially when scaling up to large, multi-story, scenes. 

Conventional data acquisition utilizes either a wheeled robot~\cite{robot_capture_2018} or human operator~\cite{backpack_2010, chen2010indoor, backpack_2012, watertight_2012, backpack_2013, watertight_2013, backpack_2015}. The former is limited in capturing all surfaces in the scene, especially in areas near the ceiling. The latter faces limitations in Simultaneous Localization and Mapping (SLAM) reconstruction due to inevitable human body pitch and roll movements~\cite{backpack_2012, song_going_2023, song_looking_2024}. Consequently, many high-fidelity systems such as Matterport~\cite{matterport} resort to a tripod-based stop-and-go capture process, which is both labor-intensive and time-consuming. Drones, however, can enable rapid and versatile capture from various perspectives, including in confined spaces. Moreover, they yield stable camera trajectories unattainable with robots or humans. Thus, this paper focuses on drone-based data capture.

Post data capture, classical approaches~\cite{schoenberger16mvs, furukawa_2010} use structure-from-motion (SfM) and multi-view stereo (MVS) to reconstruct the scene before synthesizing novel-views. 
Recently, radiance-fields methods~\cite{mildenhall2020nerf, mueller2022instantngp, Fridovich-Keil_2022_CVPR} have revolutionized the field with their photorealistic results. They implicitly represent the density and color fields of the scene and synthesize novel-views via volume rendering, allowing for view-dependent effects not easily achievable with classical approaches. Much work has been done to extend this class of methods to large scale scenes~\cite{tancik2022blocknerf,turki2022meganerf,chen2023scalarnerf}, but they mainly focus on outdoor scenes. In this paper, we develop a pipeline for large scale indoor novel-view synthesis using radiance-field approaches.  

\begin{figure}[htb]
    \centering
    \begin{subfigure}{0.4\textwidth}
        \centering
        \includegraphics[width=0.72\linewidth]{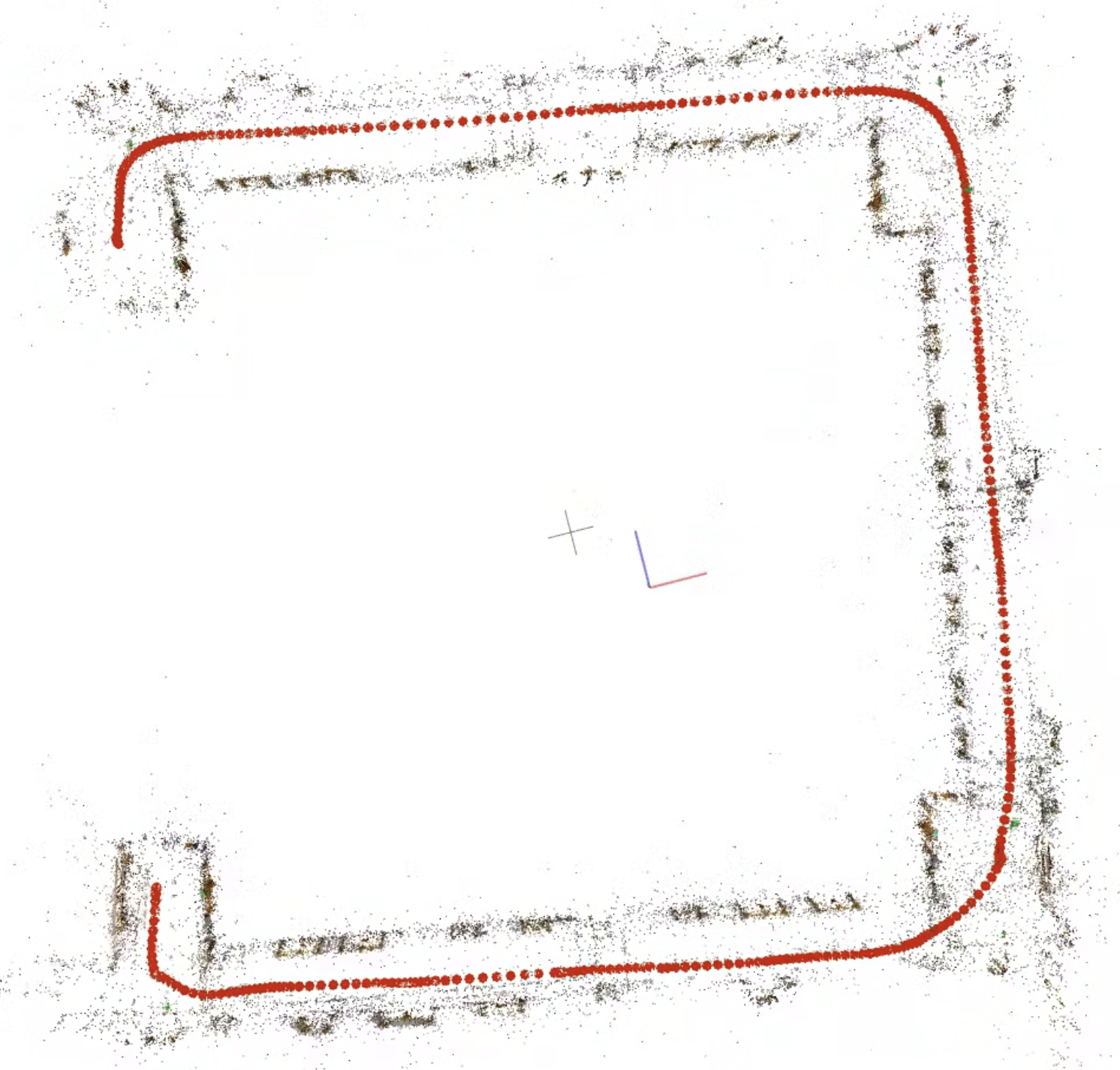}
        \caption{}
        \label{fig:drone_path_ours}
    \end{subfigure}%
    \begin{subfigure}{0.4\textwidth}
        \centering
        \includegraphics[width=0.66\linewidth]{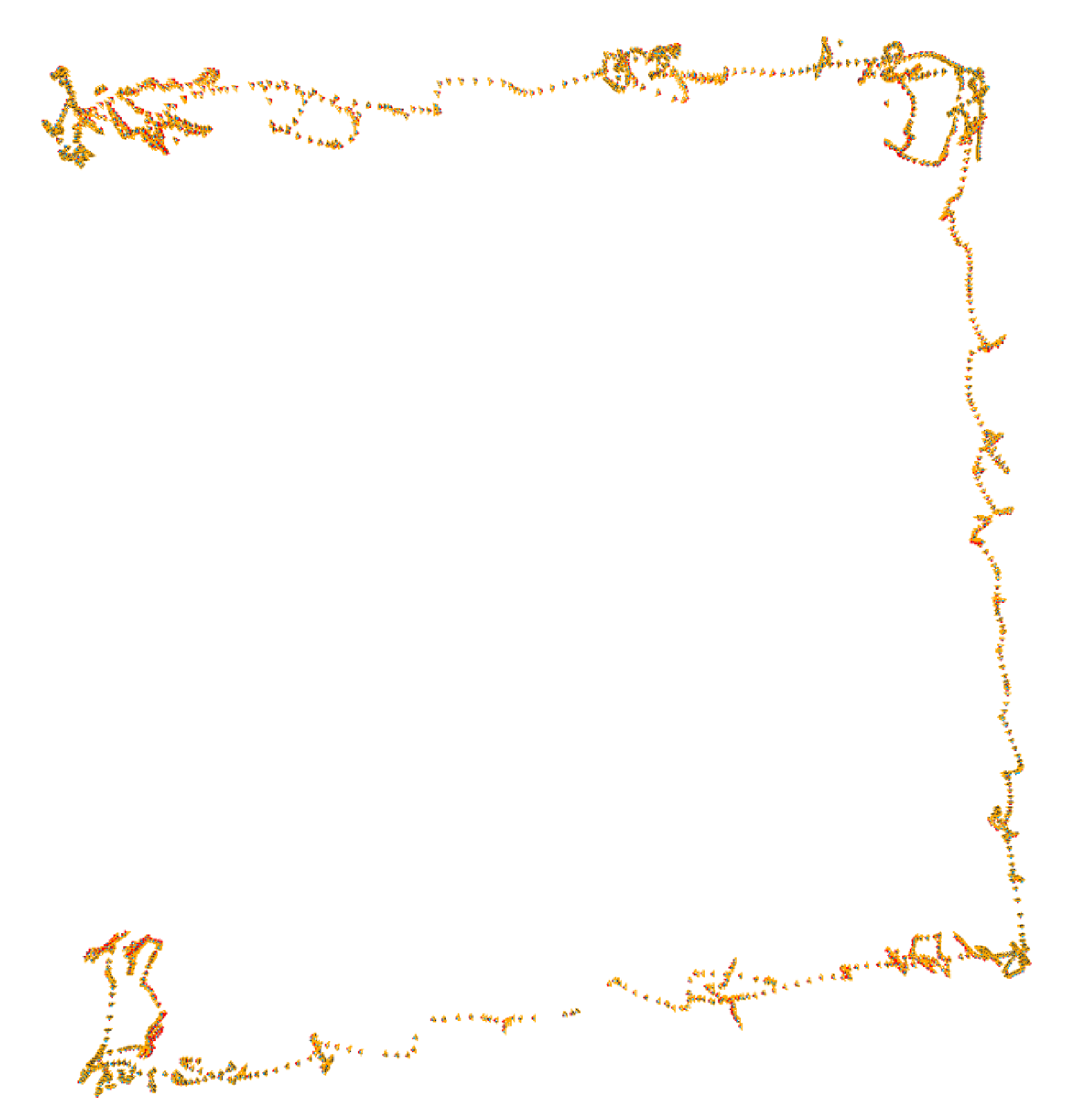}
        \caption{}
        \label{fig:drone_path_haoda}
    \end{subfigure}
    \caption{Comparison of our drone trajectory (a) and Li~\etal's~\cite{haoda_fung} (b). Our drone trajectory (in red) is straight, simple to pilot, and more reproducible than (b).}
    \label{fig:drone_paths}
\end{figure}

Recent work by Haoda~\etal~\cite{haoda_2023_ICCV, haoda_fung} propose a pipeline for large scale indoor 3D scene reconstruction from drone images. They reconstruct an explicit 3D scene representation from neural implicit surfaces under Manhattan-world assumption before synthesizing novel-views. In addition, they scale their pipeline by 
splitting up a large scene into smaller blocks. However, since they perform view synthesis after generating explicit 3D geometry, scene details and view-dependent effects are not sufficiently preserved. Furthermore, utilizing neural implicit surfaces for scene reconstruction is time consuming, taking about 5 hours for a modern machine to reconstruct a single block. Since they utilize the drone's frontal camera for data capture, they have to fly the drone in a zig-zag fashion as seen in \cref{fig:drone_path_haoda} in order to capture more diverse viewpoints. This makes drone piloting harder and the data capture process less reproducible.

We propose an indoor novel-view synthesis pipeline that is fast, photorealistic, and scalable to large complex scenes. We follow Haoda~\etal~\cite{haoda_2023_ICCV} and use a divide-and-conquer strategy to automatically split the scene into smaller blocks for both capture and processing. Rather than generating an explicit 3D structure, we use 3D Gaussian Splatting~\cite{kerbl3Dgaussians} as our scene representation and directly synthesize novel-views. In addition, we mount a 360\textdegree \,camera on top of the drone to capture a larger set of viewpoints. This allows us to fly in a simple straight drone path while capturing the scene, as seen in \cref{fig:drone_path_ours}. Furthermore, we do not employ further Laplacian filtering for blurry images like in~\cite{haoda_2023_ICCV} since our stable drone trajectory creates less motion blur.
We also develop a system that loads each reconstructed block into memory only when necessary, reducing the computational complexity of scene loading. We utilize a simple coarse-to-fine alignment method to align blocks together during the rendering stage. Our pipeline is evaluated on a large complex indoor scene with multiple rooms and corridors using a commercially available drone. Our results demonstrate marked improvement over~\cite{haoda_2023_ICCV}'s approach on the same scene by as much as $13.88 dB$ in PSNR and $0.22$ in SSIM, while achieving this at 10 times faster processing times.

\section{Related Work}
\label{sec:related_works}

\subsection{Large-Scale Scene Reconstruction with Radiance Fields}

Intrinsic limitations of NeRF \cite{mildenhall2020nerf} and its long, memory-intensive process has resulted in a variety of approaches for large-scale scene reconstruction~\cite{gu2023ue4nerfneural, tancik2022blocknerf, turki2022meganerf, chen2023scalarnerf}. In order to address model capacity issues, pipelines such as Block-NeRF separate a scene into blocks such that the NeRFs fit into memory. Scalable pipelines have also become dynamic, allowing users to update individual sections of a large-scale scene without retraining \cite{tancik2022blocknerf}. 
In general, the way blocks are split in a large scale scenes is highly scene-specific. Some scalable pipelines such as Mega-NeRF utilize a geometric clustering algorithm that allows data to be parallelized \cite{turki2022meganerf} while others such as SCALAR-NeRF utilize KMeans \cite{chen2023scalarnerf}.

\subsection{Explicit Point-Based Representation: 3D Gaussian Splatting}

Recently, explicit point-based scene representations~\cite{yifan2019DSS, aliev2020npbg, ruckert2022adop} provide a plausible approach to solve the limitations of scalable pipelines. Point-based methods explicitly represent the scene with discrete and unstructured points, usually larger than a pixel. Coupled with differentiable point-based rendering techniques \cite{wiles2020synsin, yifan2019DSS} and neural features \cite{aliev2020npbg, ruckert2022adop}, point-based methods optimize the position and opacity of points and perform fast or even real-time optimization and rendering. Nevertheless, they have poor performance in featureless or shiny areas. One seminal work that tackles both problems is 3D Gaussian Splatting \cite{kerbl3Dgaussians}. By representing the scene with more flexible 3D Gaussians, it optimizes the scene representation with adaptive density control to prevent over- or under- reconstruction of regions. With its superior performance, we base our work on 3D Gaussian Splatting for indoor scene reconstruction as well as scaling it up to large scale data.

\section{Method}
\label{sec:method}

\begin{figure}[htp]
    \centering
    \includegraphics[width=0.8\textwidth]{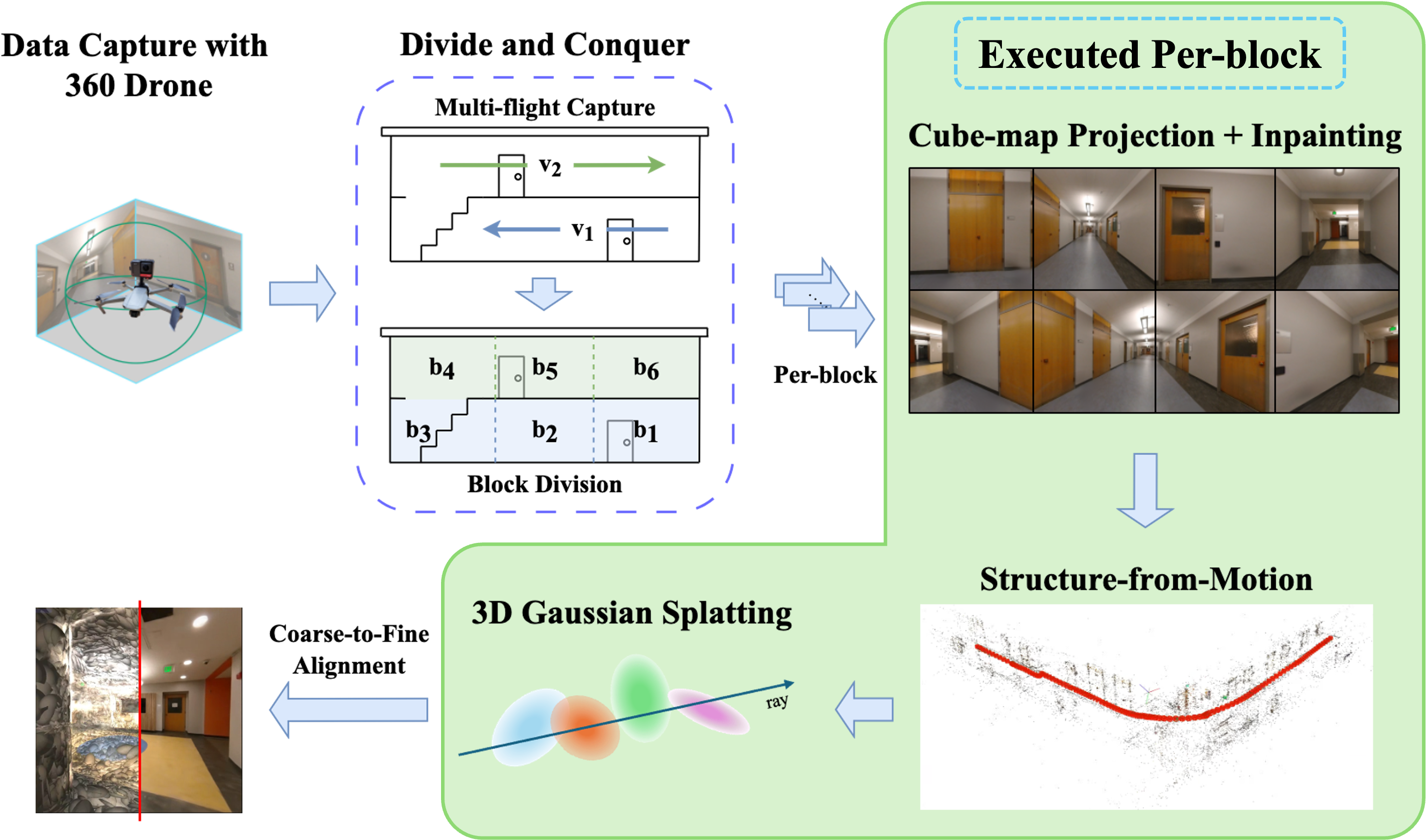}
    \caption{Our overall pipeline.}
    \label{fig:pipeline}
\end{figure}

In this section we elaborate on our data capture and view synthesis pipeline. Our pipeline takes in 360\textdegree \,video sequences captured indoor from one or multiple drone flights and outputs 3D splats as seen in \cref{fig:pipeline}. From the single or multiple flight captures, we optionally further subdivide the captured video into multiple smaller blocks. Then, we unwarp the captured spherical images via cube-mapping. This allows the images to be used further down the pipeline during SfM and 3D Gaussian Splatting. Next, we inpaint these cube-mapped images to mask out regions that contain the drone body to reduce feature matching errors. Subsequently, we retrieve camera poses and the sparse point cloud of each block via SfM and reconstruct each block with 3D Gaussian Splatting. As such, a block can either correspond to a complete video from one flight capture or a subset of it. Finally, we match the blocks together with coarse-to-fine alignment and load each block only when necessary during rendering.

\subsection{Preliminary - 3D Gaussian Splatting}\label{sec:3dgs}

3D Gaussian Splatting (3DGS) is a recent method for novel-view synthesis by Kerbl~\etal~\cite{kerbl3Dgaussians}. It represents a scene with a set of anisotropic 3D Gaussians, where each Gaussian is defined by its center position $\pmb{\mu}$ and a covariance matrix $\mathbf{\sum}$. The covariance matrix $\mathbf{\sum}$ is parameterized by a scale vector $\mathbf{s} \in \mathbb{R}^3$ and a quaternion $\mathbf{q} \in \mathbb{R}^4$ that encodes rotation of the 3D Gaussian. In addition, each 3D Gaussian is associated with an opacity $\alpha$ and a set of spherical harmonics that encodes its view-dependent color. 

As with other radiance field-based methods, SfM is required to retrieve camera poses associated with each input image. 3DGS takes in an additional sparse point cloud output from the SfM step as initialization. After training, the output 3D Gaussian Splat can be seen as a point cloud with additional effects such as anisotropic scaling, alpha-blending, and view-dependent color effects. This allows for much simpler post-processing manipulations such as our proposed coarse-to-fine point cloud alignment compared to traditional NeRF-related methods.

\subsection{Processing 360\textdegree\,Images}\label{sec:cubemap}

\begin{figure}[htp]
    \centering
    \includegraphics[width=0.65\linewidth]{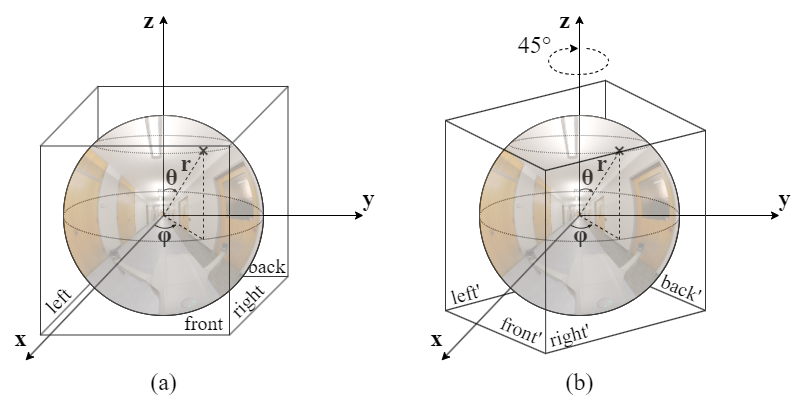}
    \caption{(a) cube-map projection. (b) 45\textdegree\,rotated projection.}
    \label{fig:projection}
\end{figure}

To facilitate SfM, we transform spherical 360\textdegree \,images into cube-map images, comprising six normal view cube faces. Mathematically, a 360\textdegree \,image sphere can be represented in polar coordinates: 
\begin{equation}\label{polar_sphere}
    x = r \sin \theta \cos \phi, \,\, 
    y = r \sin \theta \sin \phi, \,\,
    z = r \cos \theta,
\end{equation}
with $r=1$, $\theta \in [0, \pi]$, $\phi \in[0, 2\pi]$. As shown in \cref{fig:projection}(a), placing a cube with side length 2 at the origin for projection, each side face corresponds to one of four evenly divided values of $\phi$.\footnote{\url{https://stackoverflow.com/questions/29678510/convert-21-equirectangular-panorama-to-cube-map}} We use the four side faces front, left, back, right, in subsequent view synthesis. To provide sufficient number of matching features for SfM convergence, we further introduce overlapping cube-map images by rotating the 360 \textdegree \,image along the z-axis by 45\textdegree \,to obtain four additional cube faces overlapping the original four, as demonstrated in \cref{fig:projection}(b).

\subsection{Drone Body Inpainting}\label{sec:inpainting}

The eight cube-map images contain static drone body across all frames, as there is no relative motion from the 360\textdegree \,camera. This hinders feature matching of SfM, leading to poor reconstruction. One solution is to lift the camera higher so that the drone does not appear in the side cube-map images. However, this drastically changes the drone's center of gravity, leading to unstable flights. Therefore, we propose an alternative, \ie before feeding the cube-map images to SfM, we introduce an inpainting network to remove the drone body from the images. 

\begin{figure}[htp]
    \centering
    \captionsetup{justification=centering}
    \begin{tabular}{@{} >{\centering\arraybackslash} m{5.2cm} 
                    @{} >{\centering\arraybackslash} m{0.4cm} 
                    @{} >{\centering\arraybackslash} m{0.4cm} 
                    @{} >{\centering\arraybackslash} m{5.2cm} 
                    @{} >{\centering\arraybackslash} m{0.4cm} @{}}
        \includegraphics[width=0.3\linewidth]{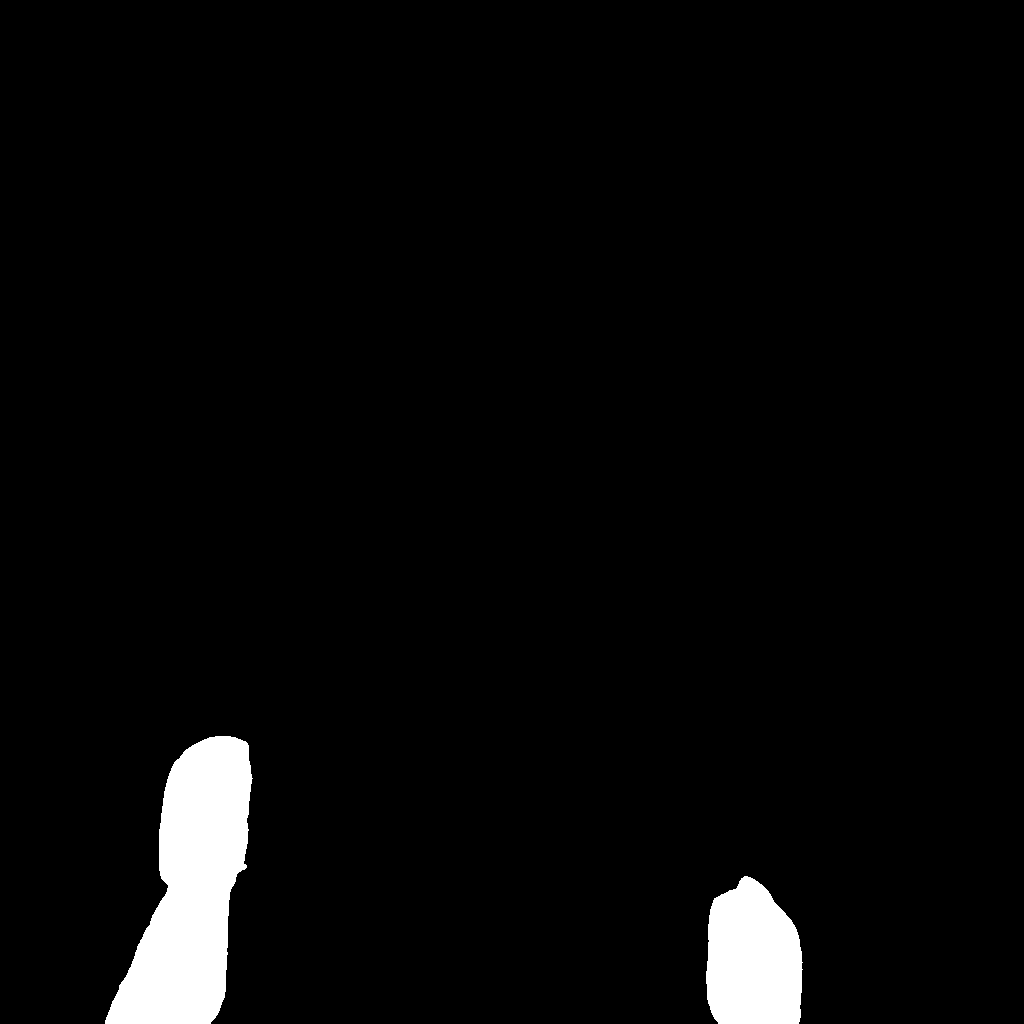}
        \includegraphics[width=0.3\linewidth]{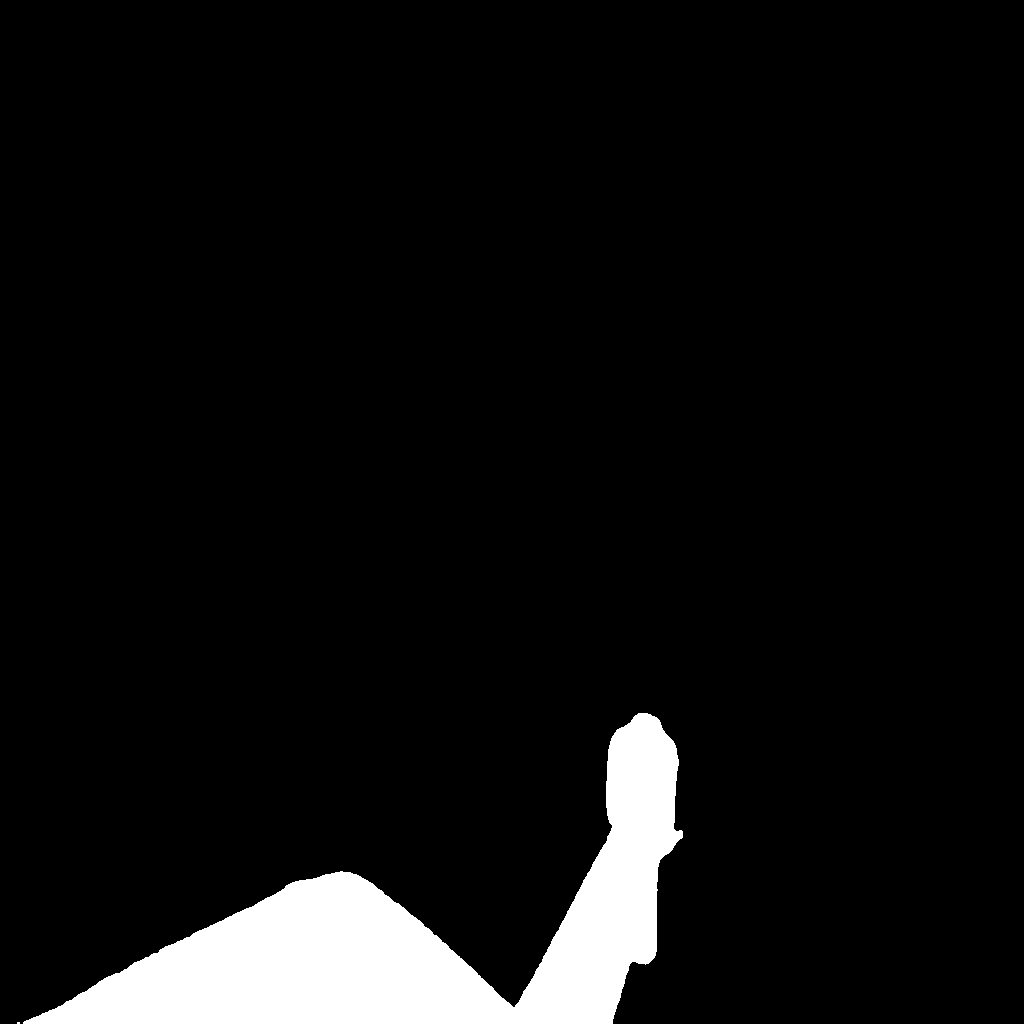}
        \includegraphics[width=0.3\linewidth]{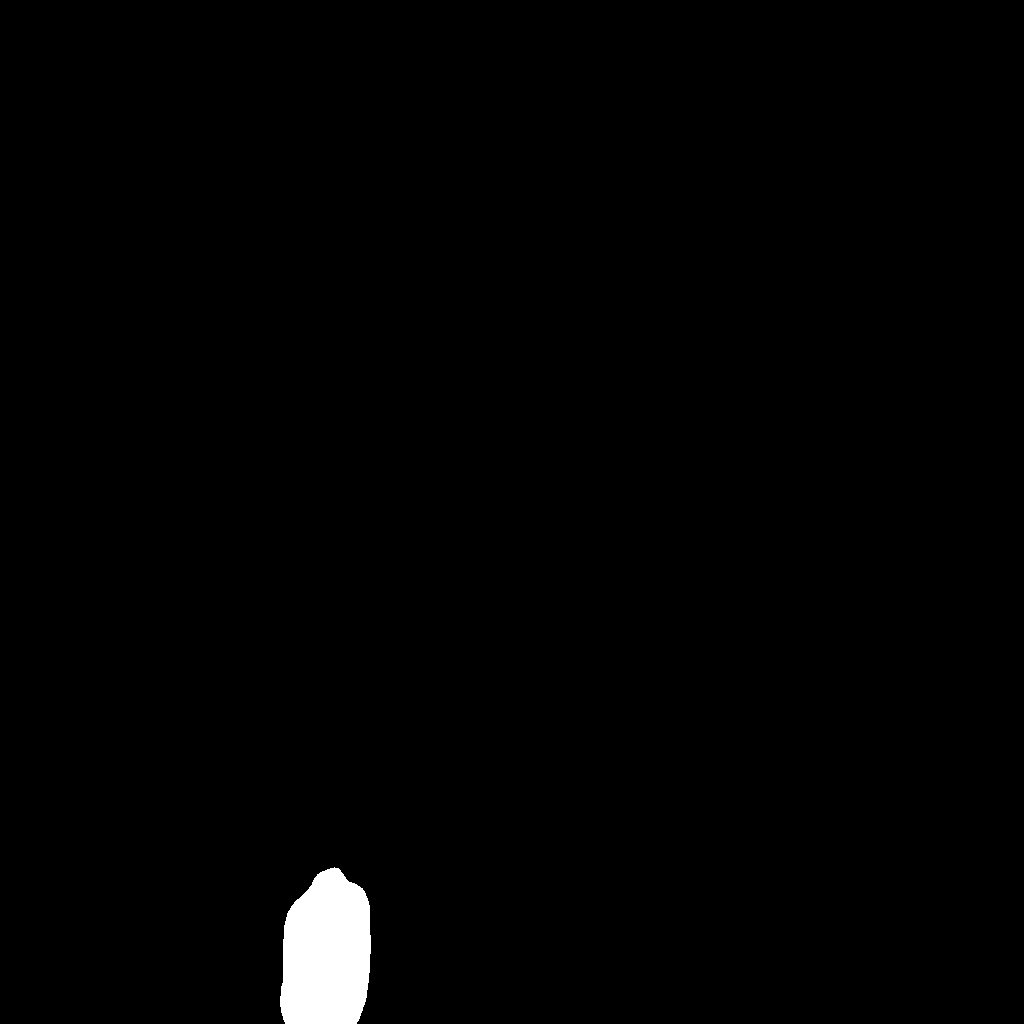}
        &$\dots$
        & \textrightarrow & 
        \includegraphics[width=0.3\linewidth]{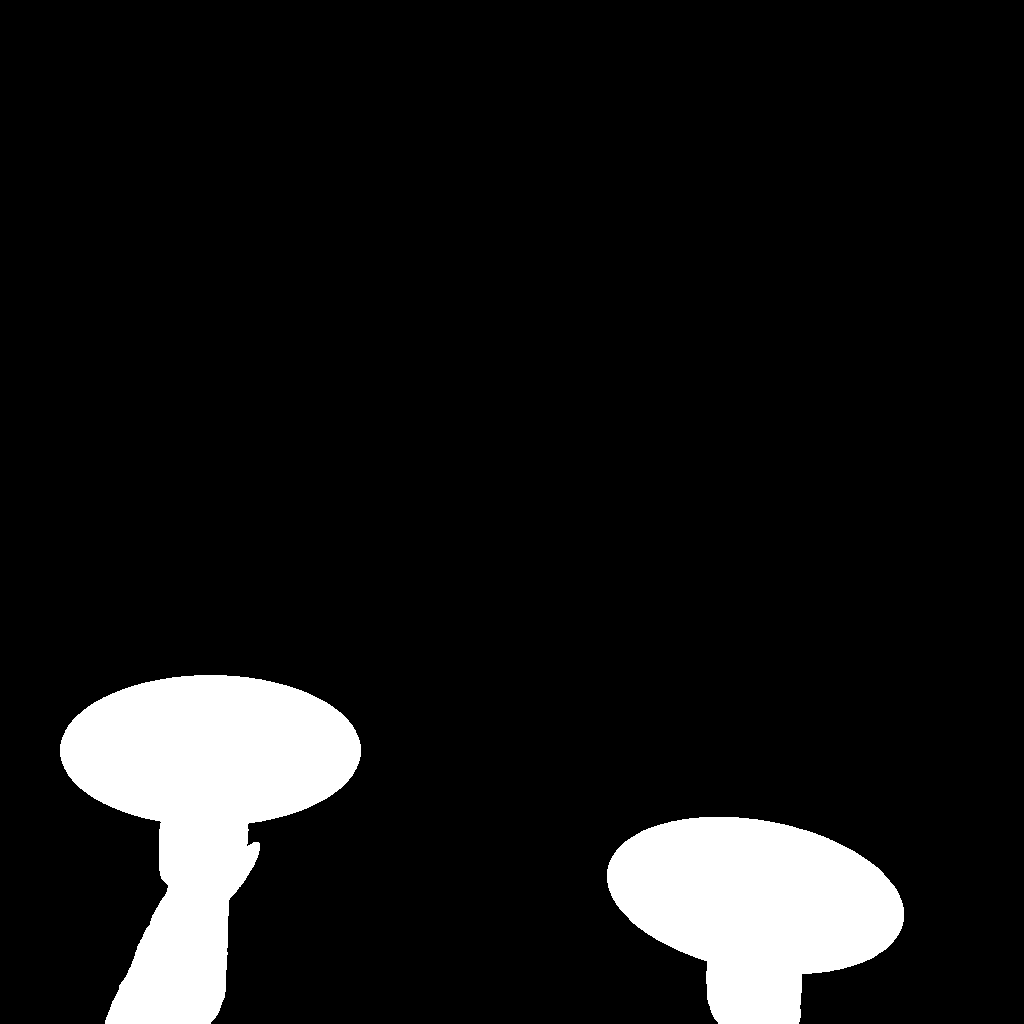}
        \includegraphics[width=0.3\linewidth]{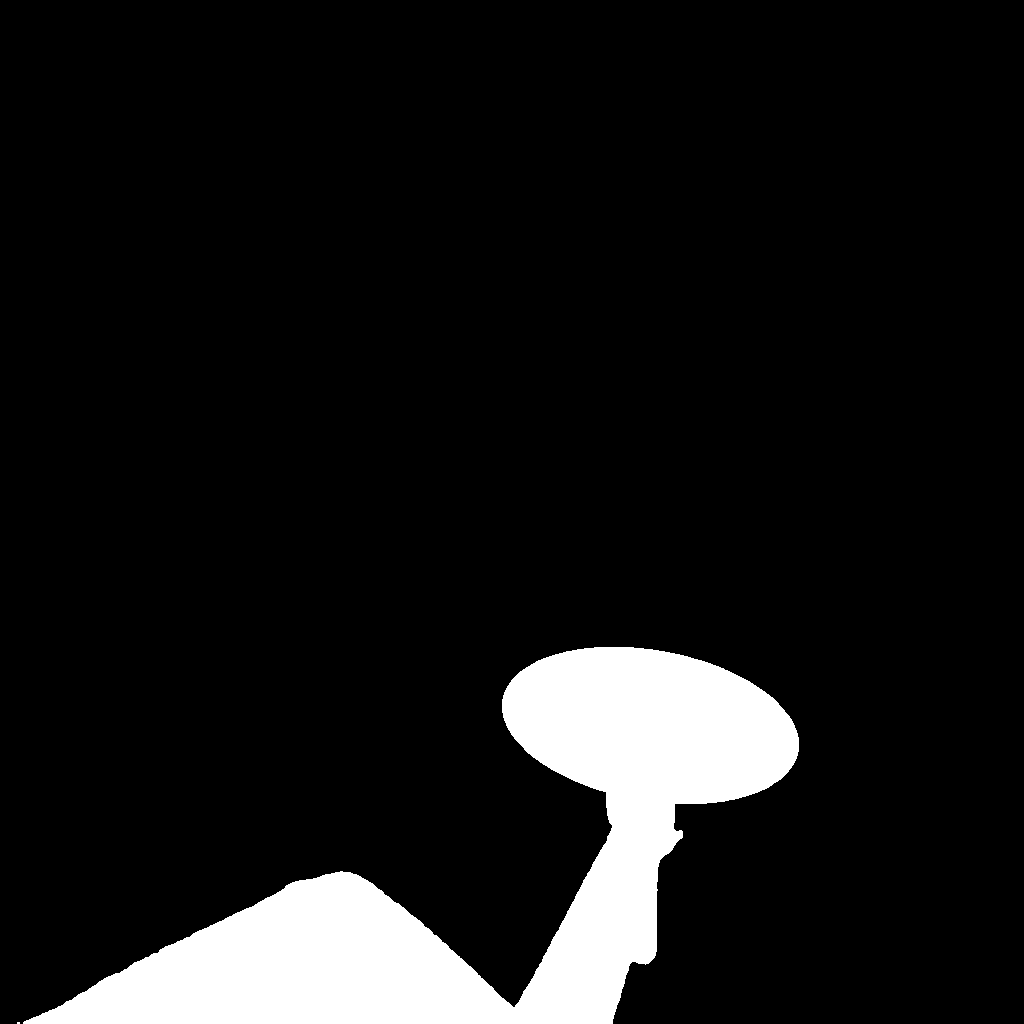}
        \includegraphics[width=0.3\linewidth]{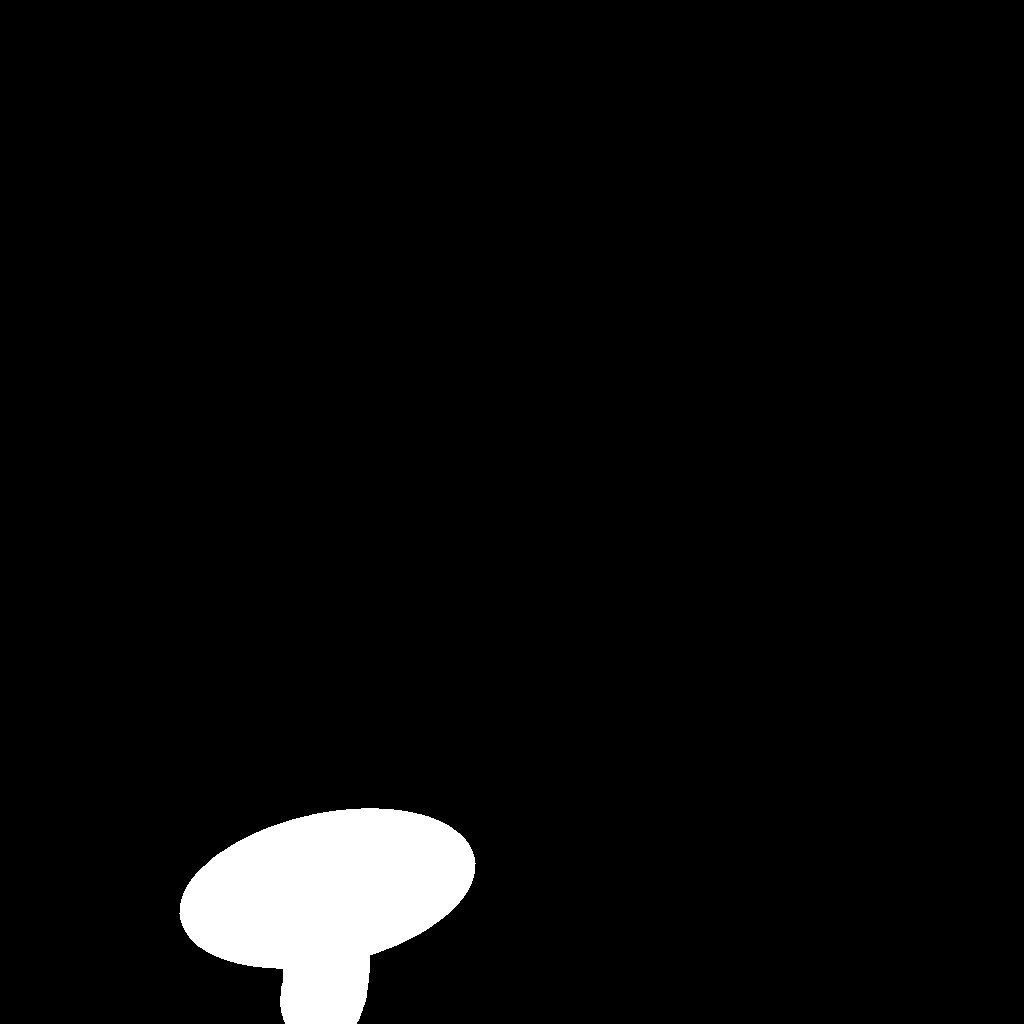}
        &$\dots$ \\
        (a) & & & (b) &\\
    \end{tabular}
    \caption{Left: SAM generated drone-body masks. Right: after adding propeller masks.}
    \label{fig:masks}
\end{figure}

We first utilize Segment Anything (SAM) \cite{kirillov2023segany} to generate accurate drone body segmentation masks for the eight cube-map images as shown in \cref{fig:masks}(a). We also find it beneficial to add elliptical masks to the tip of drone arms corresponding to the rotating propellers, shown in \cref{fig:masks}(b). To accommodate potential displacement of the mounted 360\textdegree\,camera during and across data captures, we dilate the mask by several pixels. Lastly, we employ a novel mask-guided inpainting model based on optical flow and spatiotemporal information propagation, namely, ProPainter \cite{zhou2023propainter} to inpaint out the drone whilst ensuring multi-view consistency. In our experiments, we also evaluated our pipeline with only masks without inpainting, and the results were worse.

\subsection{Divide-And-Conquer}
\label{sec:divide_and_conquer}
Novel-view synthesis on large-scale scenes is computationally expensive due to the large number of images. Moreover, large-scale indoor environments often have repetitive features, such as hallways, leading to matching errors and poor reconstruction. We thus use a divide-and-conquer approach (DAC). This strategy allows our proposed scheme to be scalable with respect to both the number of flight captures and the number of video frames in each flight capture. 

We denote our data capture strategy and block division protocol as \textit{mSnB}, indicating $m$ captured flight sequences divided into a total of $n$ blocks, where $n \geq m$. 
When $m = 1$, data is captured in a single contiguous drone flight and divided automatically into blocks. Rather than clustering as in previous works \cite{chen2023scalarnerf,turki2022meganerf}, we take advantage of the sequential nature of video data and automatically partition the video from the single flight capture into blocks with equal number of frames. We ensure 25\% overlap between adjacent blocks to reliably merge them together after DAC.
When $m > 1$, data is captured in multiple separate drone flights with an approximately 10m straight line trajectory overlap. 
Note that $m=n=1$ corresponds to the case without DAC.


DAC allows for the parallel reconstruction of each block, significantly reducing computation time. As seen later, it also mitigates erroneous feature matching when computing 3D geometries during SfM. During rendering, the blocks are loaded individually according to the viewing position. 

\subsection{Coarse-to-Fine Alignment}\label{sec:c2f_align}

Under our proposed DAC method, each block is processed separately. Despite using the same camera intrinsics across blocks, the iterative optimization in SfM results in different scales $\mathcal{S}$ for different blocks. We further need to adjust rotation matrix $\mathcal{R}$ and translation vector $\mathcal{T}$ of each block to merge them back together. 

To ensure a robust alignment, we first coarsely align $\mathcal{S}$, $\mathcal{R}$ and $\mathcal{T}$ across neighboring blocks. Given two blocks, we want to find relative scale, rotation and translation $\Delta\mathcal{S}$, $\Delta\mathcal{R}$, $\Delta\mathcal{T}$ to transform block 2 such that the overlapping section between blocks 1 and 2 are roughly aligned. 
For $\Delta\mathcal{S}$, we utilize the fact that the drone speed during data capturing is generally the same for block 1 and 2, and the frames are extracted at the same frame rate post data acquisition. We use the SfM-estimated translations $\{T_{1,i}\}_{i=1}^{M_1}$, $\{T_{2,j}\}_{j=1}^{M_2}$ for the $M_1$, $M_2$ total frames in blocks 1 and 2 respectively, and calculate $D_1$ and $D_2$, the average distance traveled in blocks 1 and 2, as in \cref{eq:coarse_align_s_1}:
\begin{equation}\label{eq:coarse_align_s_1}
\begin{aligned}
    D_1 = \frac{1}{M_1} \sum_{i=1}^{M_1-1} \left(T_{1,i+1} - T_{1,i} \right) ,\,\,
    D_2 = \frac{1}{M_2} \sum_{j=1}^{M_2-1} \left(T_{2,j+1} - T_{2,j} \right) .
\end{aligned}
\end{equation}
$D_1$ divided by $D_2$ is the $\Delta\mathcal{S}$ which we apply to block 2 for coarse scale alignment.

To obtain $\Delta\mathcal{R}$ and $\Delta\mathcal{T}$, we use the SfM-estimated camera poses $\{R_{1,i},T_{1,i}\}_{i=1}^{N_1}$ and $\{R_{2,j},T_{2,j}\}_{j=1}^{N_2}$ of the $N_1$ and $N_2$ frames in overlapping sections in blocks 1 and 2 respectively, shown in \cref{eq:coarse_align}:
\begin{equation}\label{eq:coarse_align}
    \Delta\mathcal{R} = \textit{diff}(\bar{R_{1}}, \bar{R_{2}}), \,\,
    \Delta\mathcal{T} = \textit{diff}(\bar{T_{1}}, \bar{T_{2}}),
\end{equation}
where $\bar{\cdot}$ denotes mean over set and $\textit{diff}(\cdot, \cdot)$ is the subtraction of the two poses.
We define two situations for overlapping sections.
When $m = 1$, images are from the same flight capture, the overlapping section is explicitly known and $N_1=N_2$.
When $m > 1$ \ie with multiple flight captures, we use image similarity to match the first frame of block 2, denoted by $f^{2}_{s}$ with the most similar frame $f^{1}_{s}$ in block 1, and match the last frame of block 1 $f^{1}_{e}$ with the most similar frame $f^{2}_{e}$ in block 2. We choose the most similar frame by using a simple heuristic of retrieving the image with the highest number of matching SIFT~\cite{lowe2004sift} features, explained below.

\textbf{Image Similarity with SIFT Features}
Given a query image, we want to find the most similar image $t_i$, in a sequence of ordered image frames $t_1 \dots t_i \dots t_m$, where $m$ is the total number of image frames in the comparison block. In practice, we can determine $m$ using approximate overlap region between two blocks taking into account flight capture parameters. As mentioned above, we find the most similar image by choosing the image with the highest number of matching SIFT features. To further improve robustness, we take the top-3 images with the highest number of feature matches and take the weighted average of the frame numbers by computing the softmax-probabilites using the number of feature matches. The weighted average is then rounded to the nearest integer. Since our 360\textdegree\,images generates 8 cube-mapped image frames, we perform this procedure for each side of the cube, yielding 8 numbers. Finally, we take the median of these 8 numbers to be the most similar image frame. 
 
Lastly, we use a modified version of Iterative Closest Point (ICP) \cite{besl1992icp} in CloudCompare \cite{cloudcompare} to refine the alignment between the two blocks. We use the coarse alignment result $\Delta\mathcal{S}$, $\Delta\mathcal{R}$, and $\Delta\mathcal{T}$ as the initial condition for the ICP algorithm, and iteratively refine it to obtain the final $\mathcal{S}_2$, $\mathcal{R}_2$ and $\mathcal{T}_2$.

\subsection{On-Demand Block Rendering}\label{sec:od_block_rendering}

After block alignment, we refine the functionality of the Nerfstudio~\cite{nerfstudio} interface to render blocks dynamically as required. Nerfstudio allows user to navigate freely in a scene representation, generating new views in real-time depending on the viewer's position. We adapt Nerfstudio to accommodate our large-scale scene comprising multiple blocks. 
Leveraging our DAC's block division, we optimize computation by only rendering the closest block to the viewer's position. 

\begin{figure}[htp]
    \centering
    \includegraphics[width=0.5\linewidth]{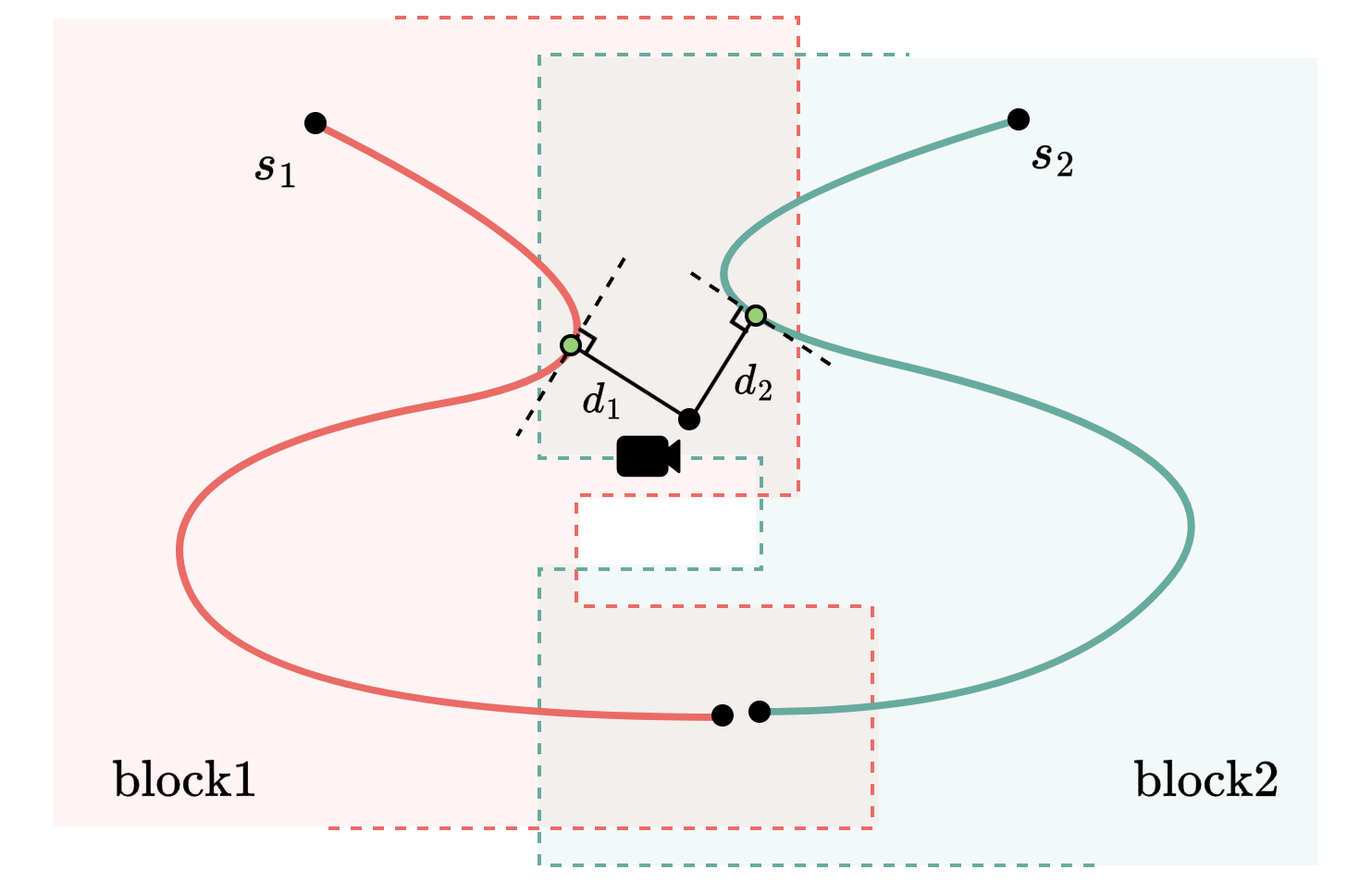}
    \caption{Block rendering protocol for complex drone trajectories.}
    \label{fig:block_loading_complex}
\end{figure}

Demonstrated in \cref{fig:block_loading_complex}, we first conduct cubic spline interpolation on the drone trajectories of blocks, where $s_1$, $s_2$, $\dots$ represent sets of piece-wise cubic polynomials with continuity $C^2$ for drone trajectories in blocks 1, 2, $\dots$. We then compute the distance $d_1$, $d_2$, $\dots$ from the viewer's position to the splines $s_1$, $s_2$, $\dots$. Finally, the block given by:  
$\argmin_{i\in\{1,2, \dots\}} d_i$
is rendered in the viewer. Note that distance calculation on splines is a minimization problem that could result in a local minima, leading to wrong block rendering. Following \cite{wang2002robust}, we use several initial estimates for BFGS-B~\cite{lbfgsb} optimization method, resulting in an accurate solution within five iterations. 

Our dynamic block rendering techniques, coupled with coarse-to-fine alignment, achieves seamless transition between blocks, as seen in our video renderings. \footnote{\url{https://youtu.be/UsZUYquSSkM} \label{note:cory}} \footnote{\url{https://youtu.be/MhVooLRnmhE} \label{note:hmmb}} Qualitative results are included in \cref{sec:results}.

\section{Experiments}
\label{sec:exps}

\subsection{Datasets and Capture Procedure}
To evaluate the effectiveness of our proposed pipeline, we captured and reconstructed two indoor scenes: (1) 3rd floor of Cory Hall, and (2) the main lobby of Hearst Memorial Mining Building. Both indoor scenes are academic buildings located in University of California, Berkeley. 

\textbf{Indoor Scene 1 - Cory Hall}
We capture this scene to compare our pipeline against Li~\etal's~\cite{haoda_2023_ICCV} pipeline on the same scene.

\begin{figure}[htb]
    \centering
    \begin{subfigure}{0.3\textwidth}
        \centering
        \includegraphics[width=0.95\linewidth]{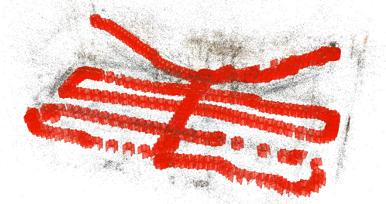}
        \caption{}
        \label{fig:hmmb_floor1}
    \end{subfigure}%
    \begin{subfigure}{0.3\textwidth}
        \centering
        \includegraphics[width=0.95\linewidth]{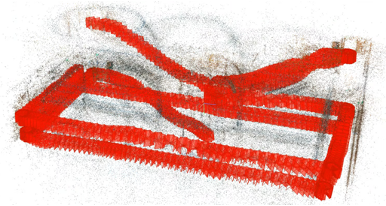}
        \caption{}
        \label{fig:hmmb_floor2}
    \end{subfigure}%
    \begin{subfigure}{0.3\textwidth}
        \centering
        \includegraphics[width=0.95\linewidth]{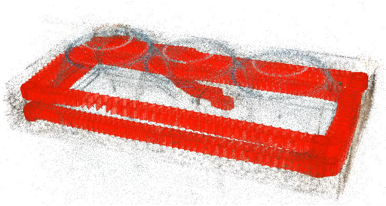}
        \caption{}
        \label{fig:hmmb_floor3}
    \end{subfigure}
    \caption{Drone trajectories for all three floors of Hearst Memorial Mining Building. (a) 1F: the two loops in front is in the empty space of the first floor and the diagonal pathways towards the back are along the staircases leading to 2F. (b) 2F: the drone flew 2 loops along the corridor and on the staircases leading from 1F and to 3F. (c) 3F: similar to 2F but only one staircase leading from 2F.}
    \label{fig:hmmb_paths}
\end{figure}

\textbf{Indoor Scene 2 - Hearst Memorial Mining Building Main Lobby}
We also captured the main lobby of the Hearst Memorial Mining Building at University of California, Berkeley, to evaluate our pipeline's effectiveness on a large, building-scale scene. This scene consists of $3$ floors with tall ceilings and multiple staircases. Due to its scale it is not feasible to capture the entire scene in a single drone flight. Thus, we separate the scene into $3$ drone flights, one flight per floor. The first floor of the scene contains a large empty space. Thus, we fly the drone in the trajectory shown in \cref{fig:hmmb_floor1} in order to fully cover the floor area. As for the second and third floors, we fly the drone only along the corridors and staircases of the scene, as seen in \cref{fig:hmmb_floor2} and \cref{fig:hmmb_floor3}. We could not fly the drone in the center as that would trigger the fire alarm of the building. Hence, we opted to fly two loops along the corridors at different heights in order to capture sufficient viewpoints of the empty atrium in the center. We ensure overlap between floors along the staircases, as shown in \cref{fig:hmmb_paths}. 


\subsection{Implementation Details}

For data acquisition of both scenes, we utilize the Insta360 ONE RS camera and mount it onto a DJI Mavic Air 2 drone to capture 360\textdegree\,videos. All videos are captured at $30$ frames per second (FPS). After data capture, we use Insta360 Studio to export the 360\textdegree\,videos into $5760 \times 2880$ equirectangular MP4 videos. We then extract equirectangular frames from these videos at $3$ FPS for scene 1 and $1$ FPS for scene 2. Subsequently, these equirectangular frames are cube-mapped to $768 \times 768$ sized images for the SfM step. We use sparse reconstruction in COLMAP \cite{schoenberger16mvs} as the SfM method. Our data processing and model training is performed on a NVIDIA TITAN RTX 24GB GPU. For 3D Gaussian Splatting, we used the original implementation proposed by Kerbl~\etal~\cite{kerbl3Dgaussians}.

\subsection{Results}\label{sec:results}

\subsubsection{Scene 1 - Cory Hall}
We compare our method with~\cite{haoda_2023_ICCV} in terms of rendered image quality and computation time. We use peak-signal-to-noise ratio (PSNR) and structural similarity index measure (SSIM) for image quality assessment.

\begin{table}[htb]
\begin{center}
    {\resizebox{\linewidth}{!}{
\begin{tabular}{|c|c|c|c|c|c|c|c|c|}
    \hline
    Method & \# Flight Captures & \# Blocks & Frame Size & \# Frames & PSNR\textuparrow & SSIM\textuparrow & Time\textdownarrow (hrs) & \# Pixels \\
    \hline\hline
    \cite{haoda_2023_ICCV}  & 1 & 5 & 1360 $\times$ 765 & 2000 & 20.76 & 0.74 & 25 & 2 B\\
    \textit{1S1B} (w/o DAC) & 1 & 1 & 768 $\times$ 768  & 2552 & 31.31 & 0.95 & 3 & 1.5 B\\
    \textbf{\textit{1S4B}}  & 1 & 4 & 768 $\times$ 768  & 2552 &  \textbf{34.03} & \textbf{0.96} & \textbf{0.5+4$\cdot$(0.5)} & \textbf{1.5 B}\\
    \textbf{\textit{4S4B}}  & 4 & 4 & 768 $\times$ 768  & 5632 &  \textbf{34.64} & \textbf{0.96} & \textbf{0.5+4$\cdot$(0.5)} & \textbf{3.3 B}\\
    \hline
\end{tabular}
}}
\end{center}
\caption{Quantitative comparison between pipelines.}
\label{tab:comp_nerf}
\end{table}

In \cref{tab:comp_nerf}, we quantitatively compare three versions of our method, \textit{1S1B}, \textit{1S4B}, \textit{4S4B} with \cite{haoda_2023_ICCV} using PSNR and SSIM.  As seen, all three versions of our approach outperform \cite{haoda_2023_ICCV} by as much as $13.88 dB$ in PSNR and $0.22$ in SSIM.
Our method \textit{1S4B} significantly outperforms \cite{haoda_2023_ICCV} with an increase of $13.3 dB$ PSNR and $0.22$ SSIM, despite the fact that its total pixel count, \ie number of pixels per frame times total number of frames, is $75\%$ of \cite{haoda_2023_ICCV}. Moreover, our method takes much less computation time overall, taking only 2.5 hours as opposed to 25 hours in~\cite{haoda_2023_ICCV}, a 10 times speed improvement. Furthermore, if parallel processing is employed, our pipeline can complete scene reconstruction in as little as 1 hour. 

As seen in \cref{tab:comp_nerf}, \textit{4S4B} achieves $0.61dB$ higher PSNR than \textit{1S4B}. This is likely due to the introduction of $50\%$ new pixels in the overlapping sections from separate drone flights. Comparing \textit{1S4B} and \textit{1S1B}, the number of pixels are the same due to the same underlying data, yet \textit{1S4B} achieves a higher PSNR and SSIM, indicating an inherent advantage of DAC. This is surprising as SfM on the whole scene without DAC is generally believed to have more constraints for optimization and should produce better results. We speculate that DAC mitigates erroneous feature matching by limiting number of repetitive/plain features. Our experiments also show a slight reduction of 0.5 hours in pipeline time utilizing DAC. More importantly, DAC allows time per-block shown in brackets in \cref{tab:comp_nerf} to be parallelized for more significant time reduction in our method.


\begin{figure}[ht]
\setlength\tabcolsep{1.2pt} 
\centering
\begin{tabular}{@{} r M{0.22\linewidth} M{0.22\linewidth} M{0.22\linewidth} M{0.22\linewidth} @{}}
\begin{subfigure}{0.05\linewidth} \caption*{Ours}\label{} \end{subfigure} 
  & \includegraphics[width=\hsize]{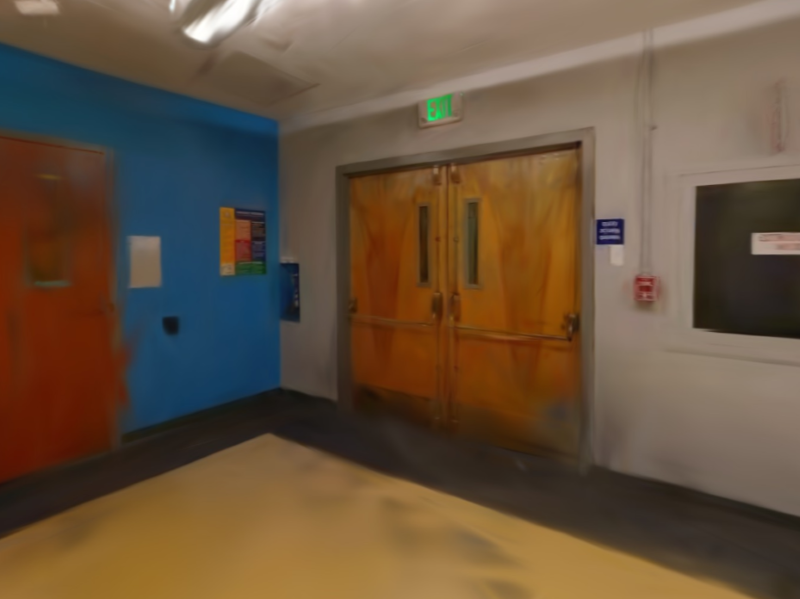} 
  & \includegraphics[width=\hsize]{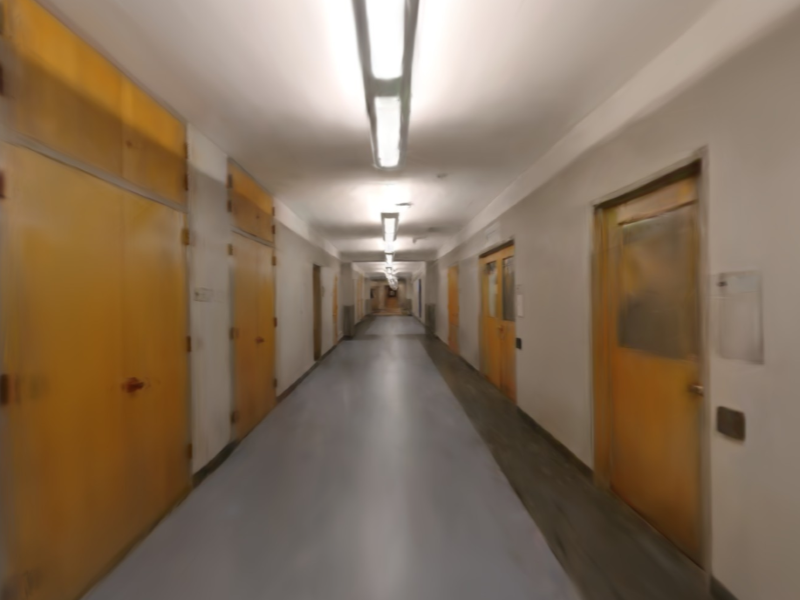}
  & \includegraphics[width=\hsize]{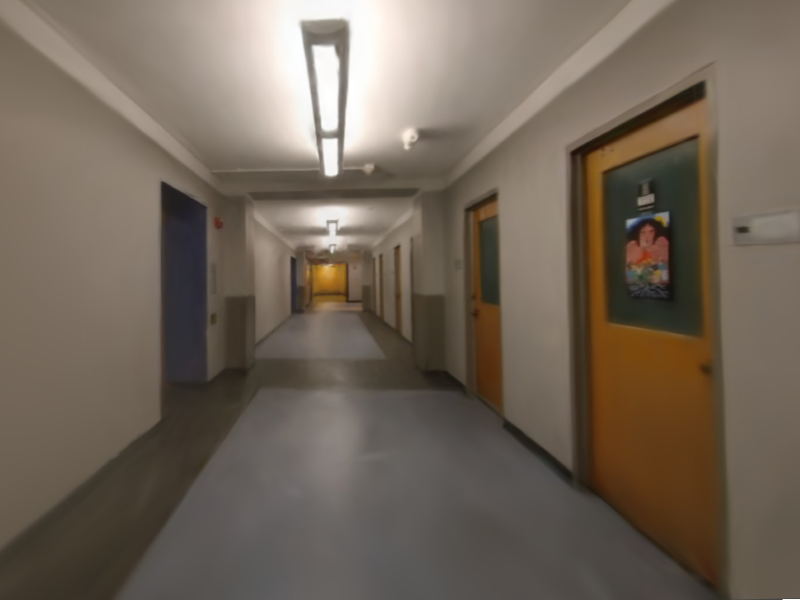}
  & \includegraphics[width=\hsize]{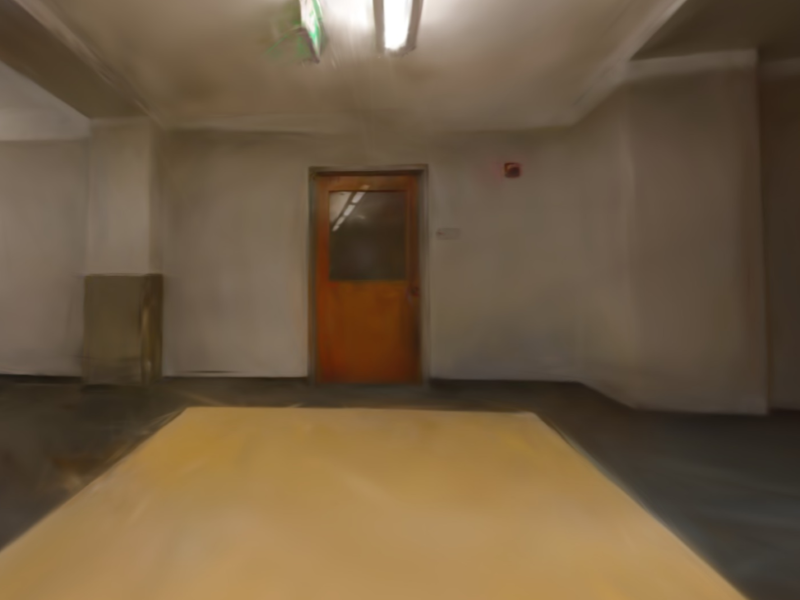}\\
\begin{subfigure}{0.05\linewidth} \caption*{\cite{haoda_2023_ICCV}}\label{} \end{subfigure} 
  & \includegraphics[width=\hsize]{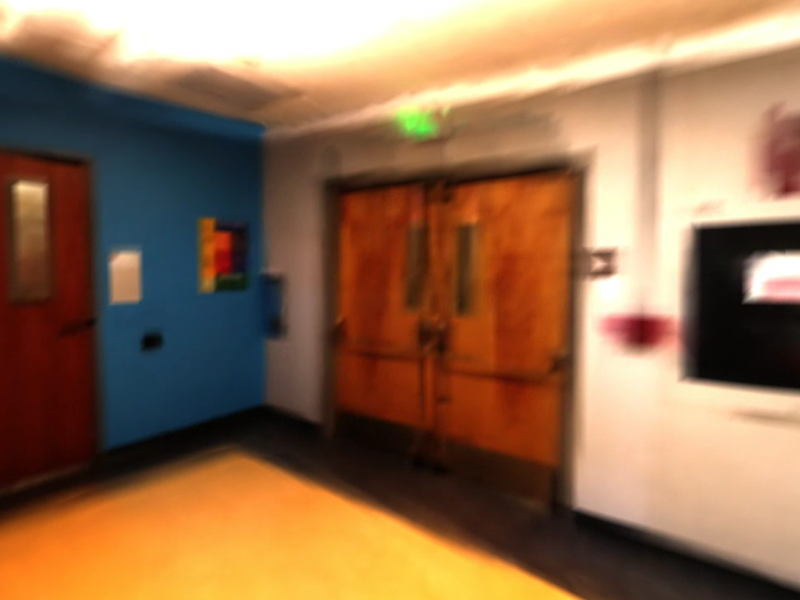}    
  & \includegraphics[width=\hsize]{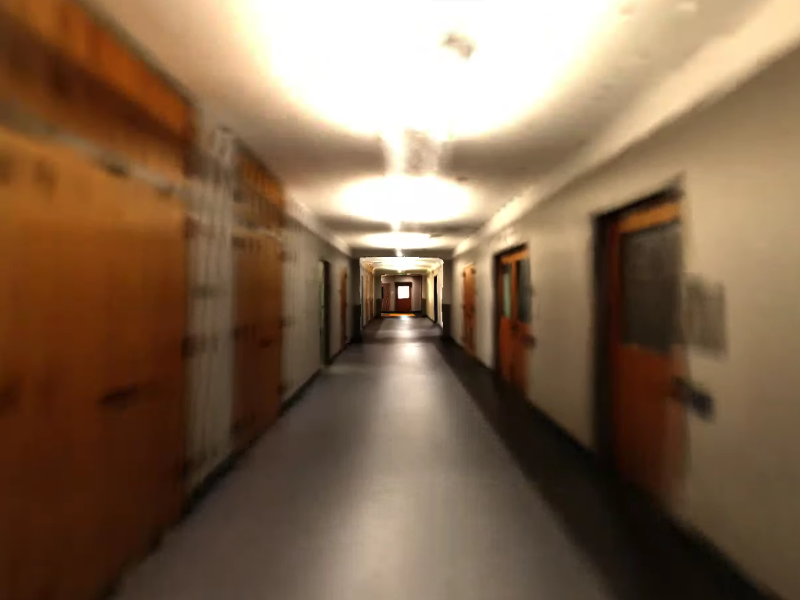}
  & \includegraphics[width=\hsize]{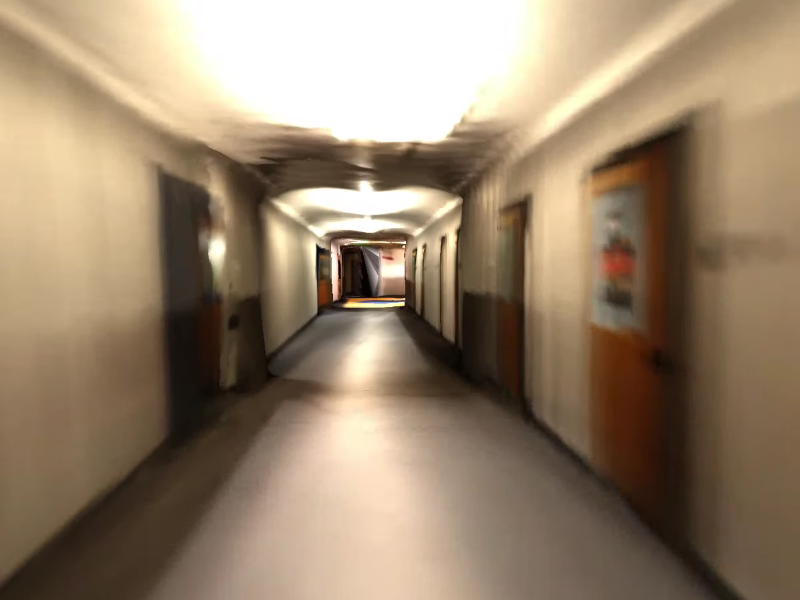}
  & \includegraphics[width=\hsize]{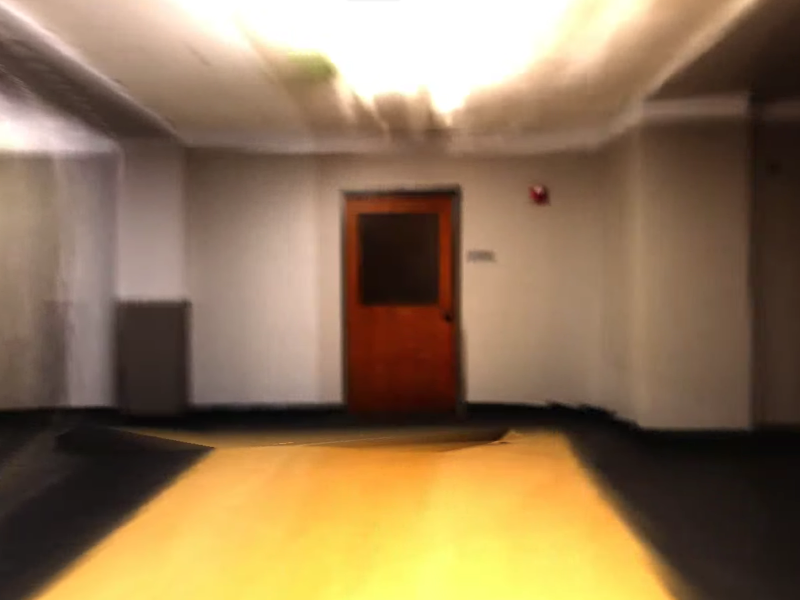}\\
\end{tabular}
\caption{Novel-view renderings of \textit{1S4B} (top) and \cite{haoda_2023_ICCV} (bottom).}
\label{fig:comp_nerf}
\end{figure}

A more qualitative comparison between our \textit{1S4B} and \cite{haoda_2023_ICCV} can be seen in \cref{fig:comp_nerf}. Column 1 shows that our method results in much sharper reconstruction than~\cite{haoda_2023_ICCV}. In corridors shown in column 2, \textit{1S4B} generates better details on walls even though they are parallel to the drone trajectory. We attribute this to our use of the 360\textdegree\,camera, which captures more viewpoints in a single forward flight. The image quality improvement from utilizing the 360\textdegree\,camera can also be seen in column 3 when rendering ``backwards'' relative to the drone trajectory. Lastly, as seen in column 4, the use of 3DGS allows dynamic reflective surfaces.

\begin{figure}[htb]
\setlength\tabcolsep{1pt} 
\centering
\begin{tabular}{@{} r M{0.3\linewidth} M{0.3\linewidth} M{0.3\linewidth} @{}}
& \textit{1S1B} & \textit{1S4B} & \textit{4S4B} \\ \addlinespace
\begin{subfigure}{0.05\linewidth} \caption*{$T-1$}\label{} \end{subfigure} 
  & \includegraphics[width=\hsize]{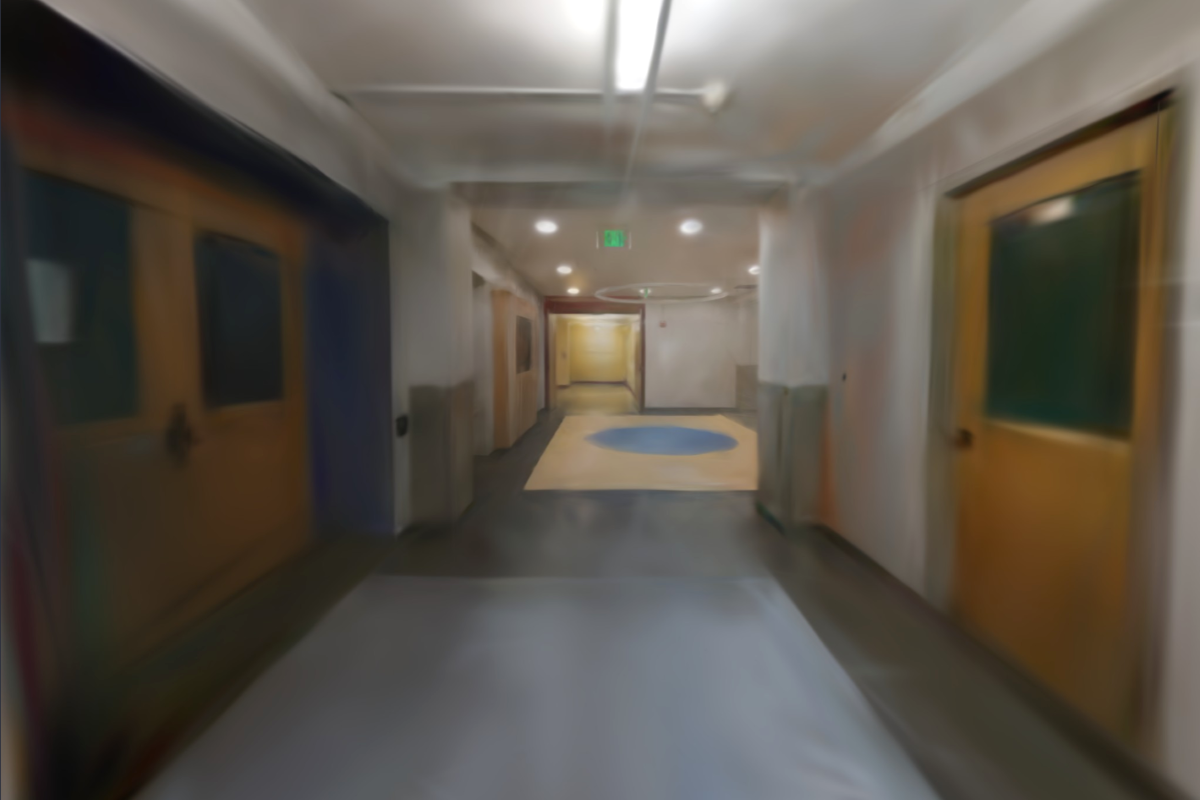} 
  & \includegraphics[width=\hsize]{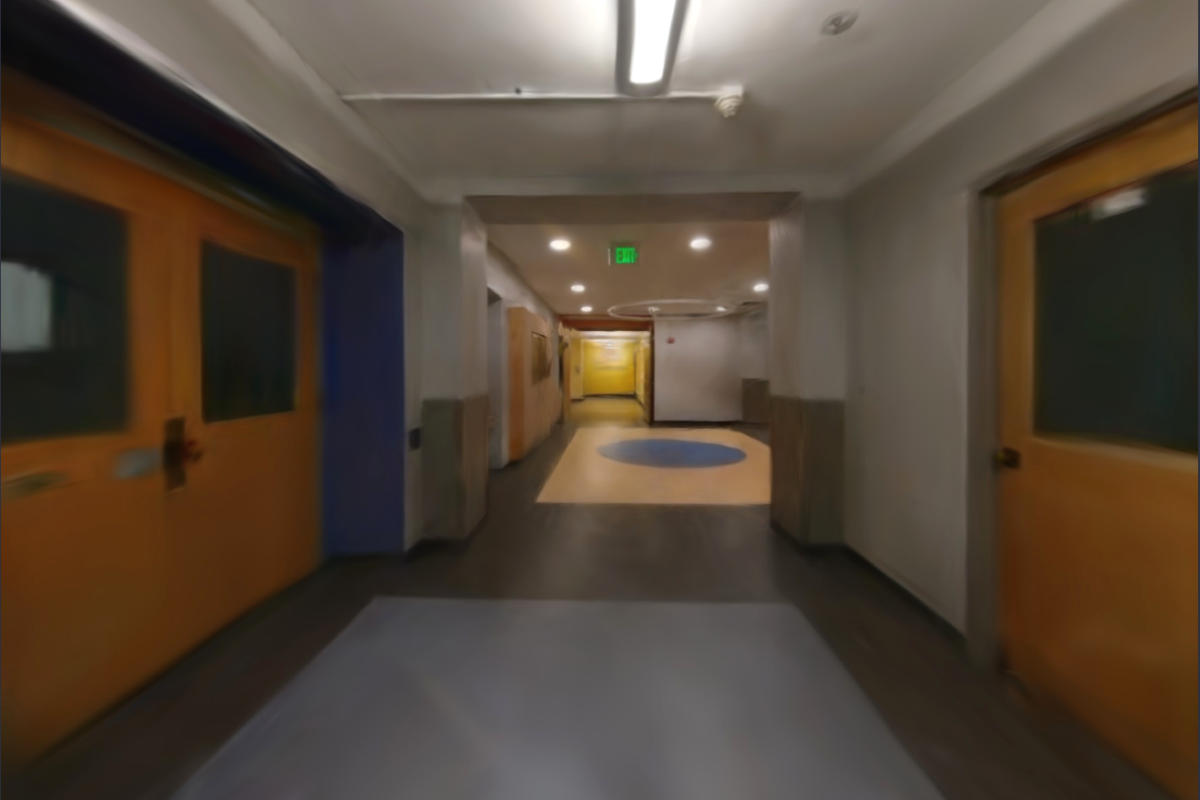}
  & \includegraphics[width=\hsize]{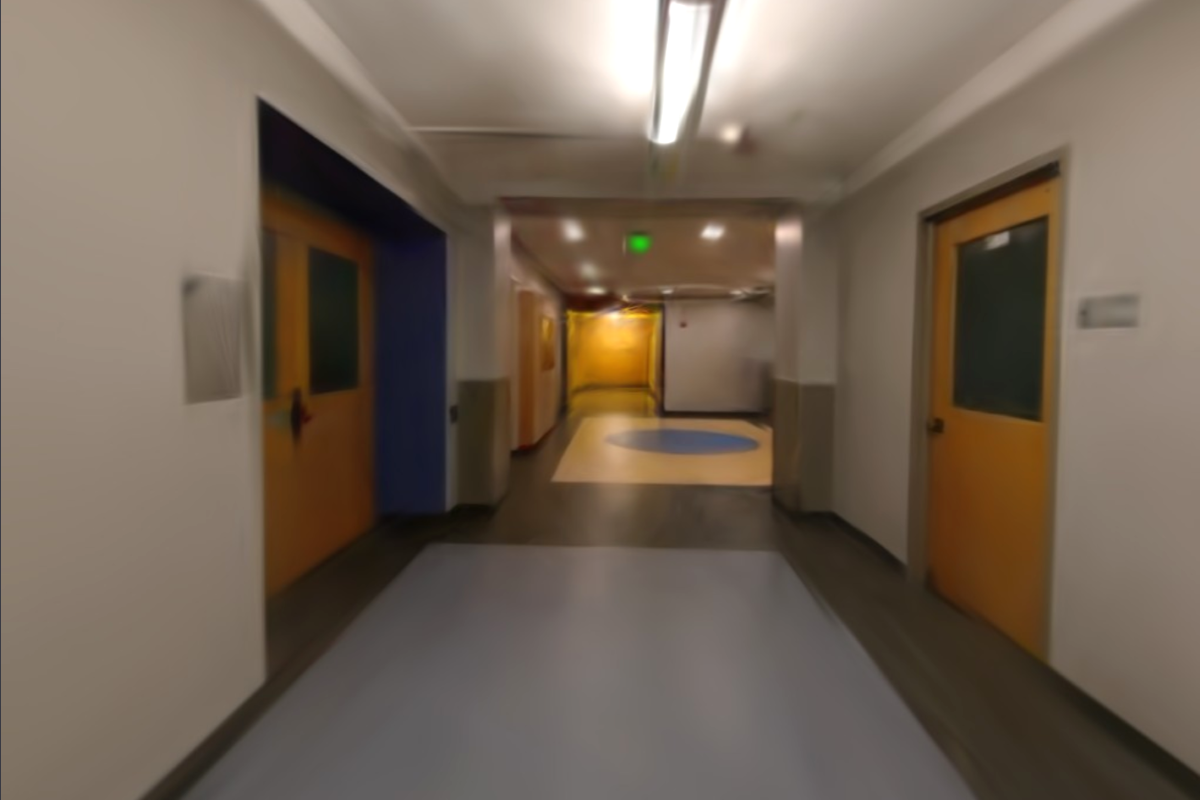}\\
\begin{subfigure}{0.05\linewidth} \caption*{$T$}\label{} \end{subfigure} 
  & \includegraphics[width=\hsize]{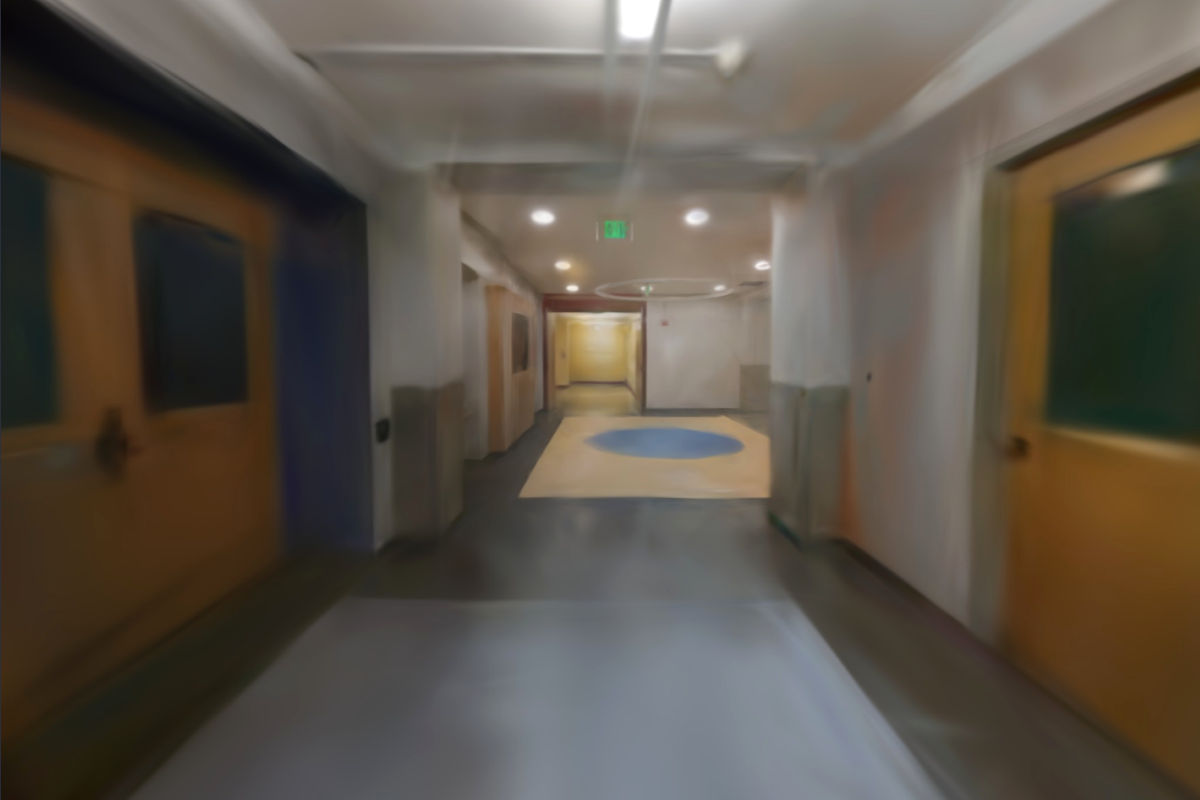}    
  & \includegraphics[width=\hsize]{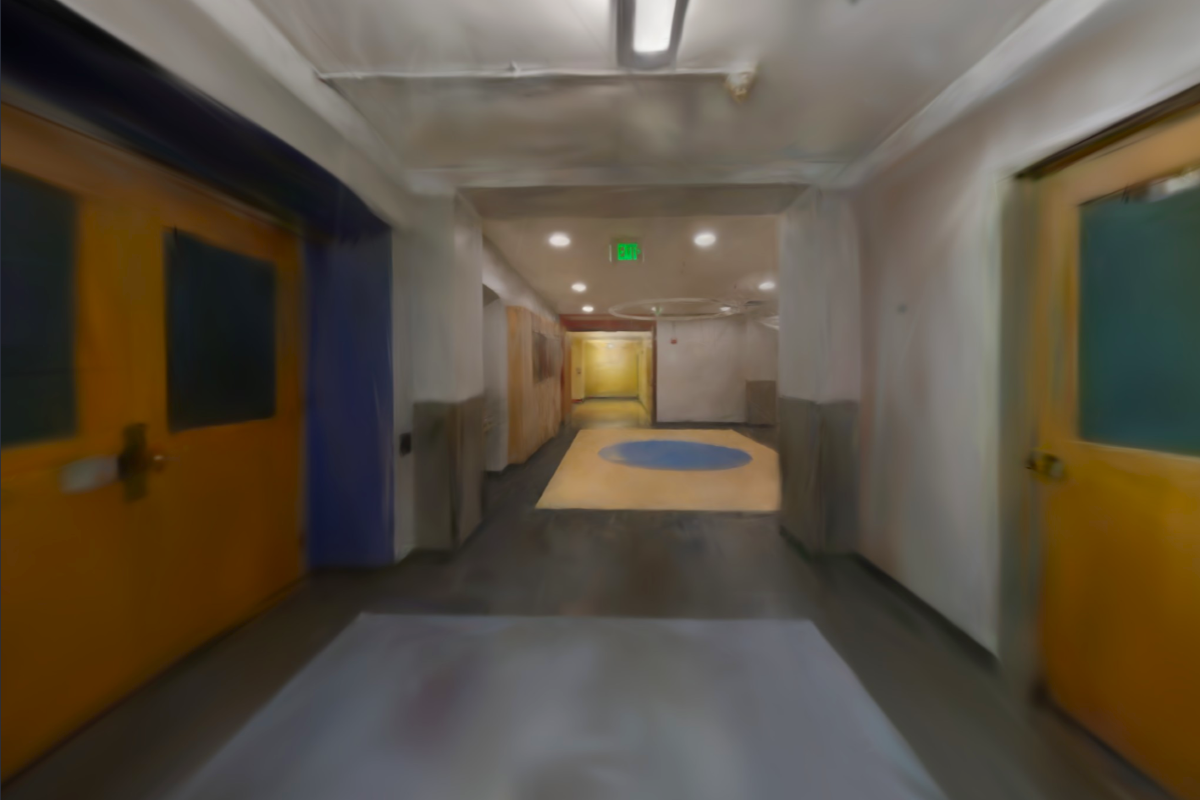}
  & \includegraphics[width=\hsize]{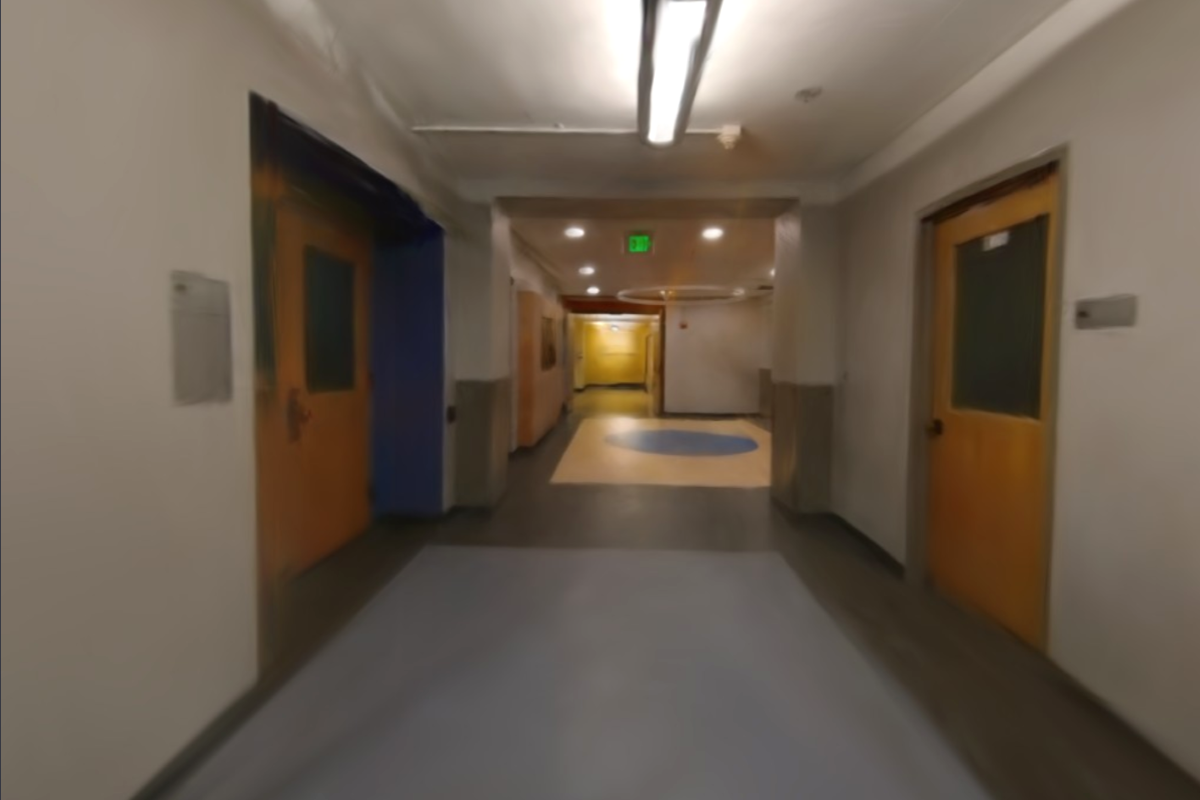}\\
\end{tabular}
\caption{Consecutive renderings at timestamps \textbf{T - 1} and \textbf{T} at block boundary with different data capture protocols. Note \textit{1S4B} and \textit{4S4B} have different block boundary since \textit{1S4B} is automatically divided.}
\label{fig:comp_3methods}
\end{figure}

\cref{fig:comp_3methods} shows consecutive renderings at block boundary with different methods. With \textit{1S4B} and \textit{4S4B}, the transition between two blocks is smooth, thanks to our coarse-to-fine alignment. Moreover, the DAC rendering quality in \textit{1S4B} and \textit{4S4B} is visually superior to \textit{1S1B}. \textit{4S4B} video can be found at \cref{note:cory}.

\begin{table}[htb]
\begin{center}
\begin{tabular}{|c|c|c|c|c|}
    \hline
    Method & PSNR\textuparrow & SSIM\textuparrow & Time\textdownarrow & \# Pixels \\
    \hline\hline
     \textit{3S3B} & 27.74 & 0.89 & 1+3$\cdot$(6) hrs & 7.1 B\\
    \hline
\end{tabular}
\end{center}
\caption{Quantitative results on Hearst Memorial Mining Building.}
\label{tab:hmmb}
\end{table}

\subsubsection{Scene 2 - Hearst Memorial Mining Building (HMMB)}

\begin{figure}[htb]
\setlength\tabcolsep{1.2pt} 
\centering
\begin{tabular}{@{} r M{0.2\linewidth} M{0.2\linewidth} M{0.2\linewidth} M{0.2\linewidth} @{}}
\begin{subfigure}{0.05\linewidth} \end{subfigure} 
  & \includegraphics[width=\hsize]{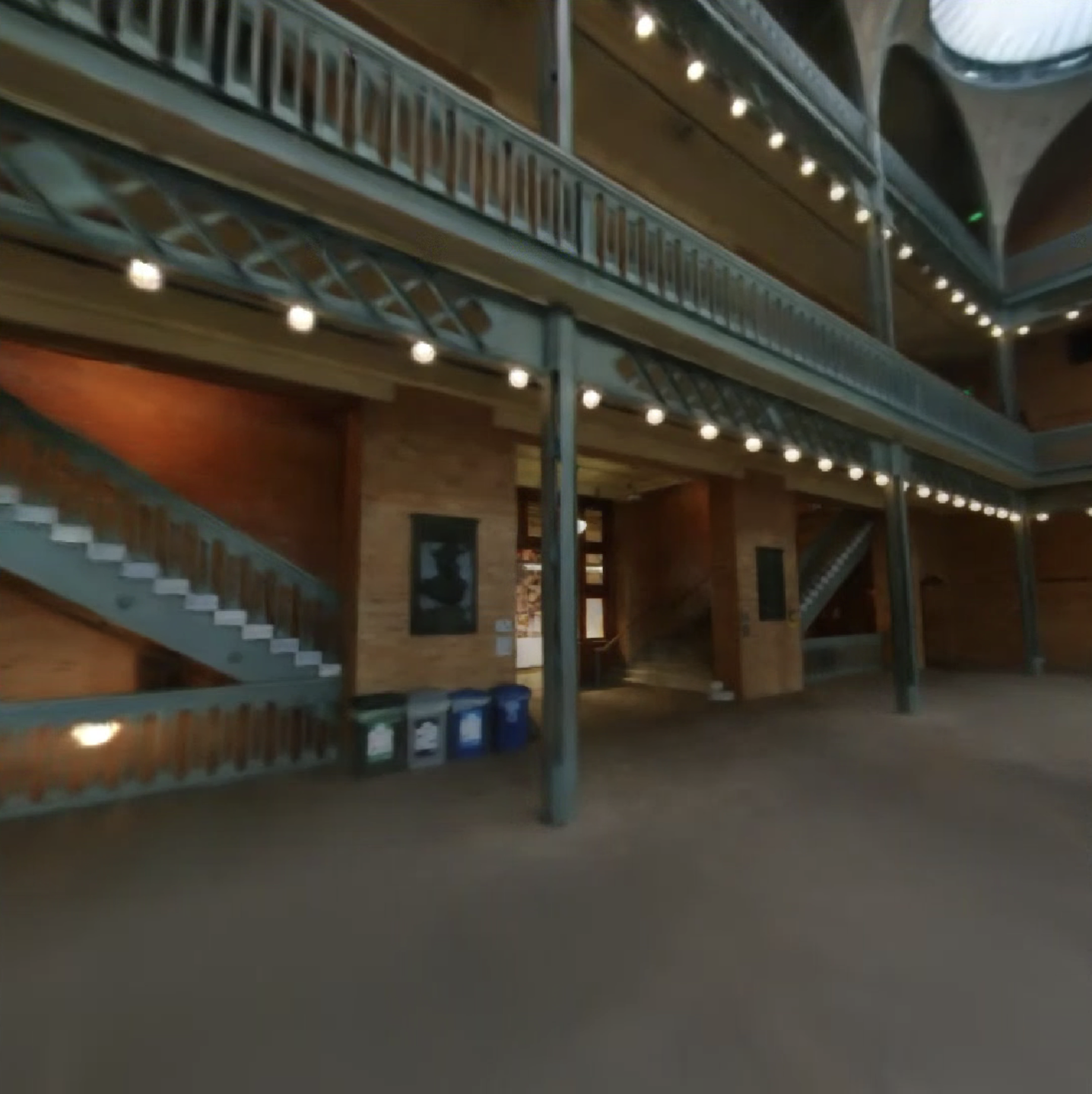} 
  & \includegraphics[width=\hsize]{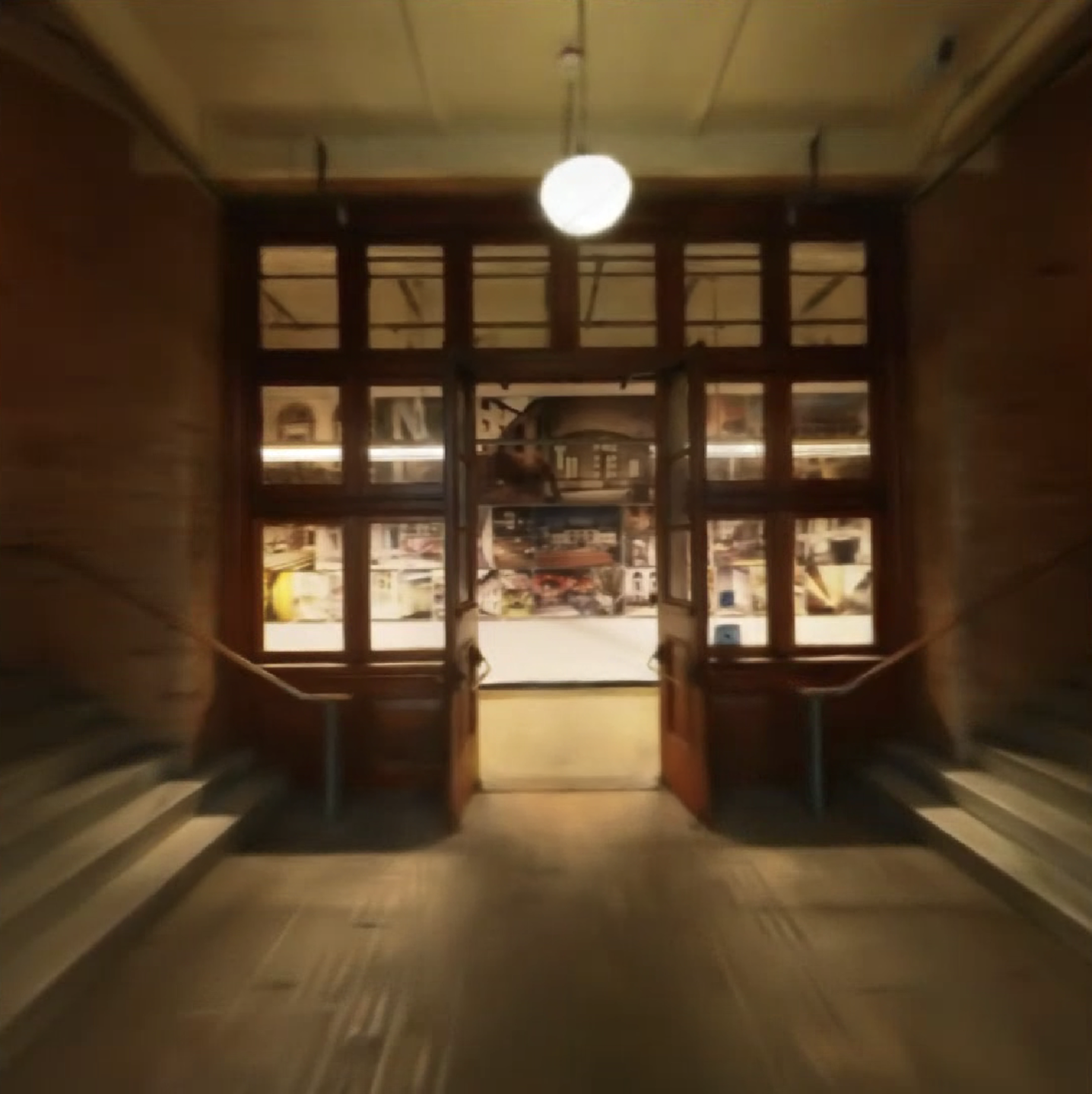}
  & \includegraphics[width=\hsize]{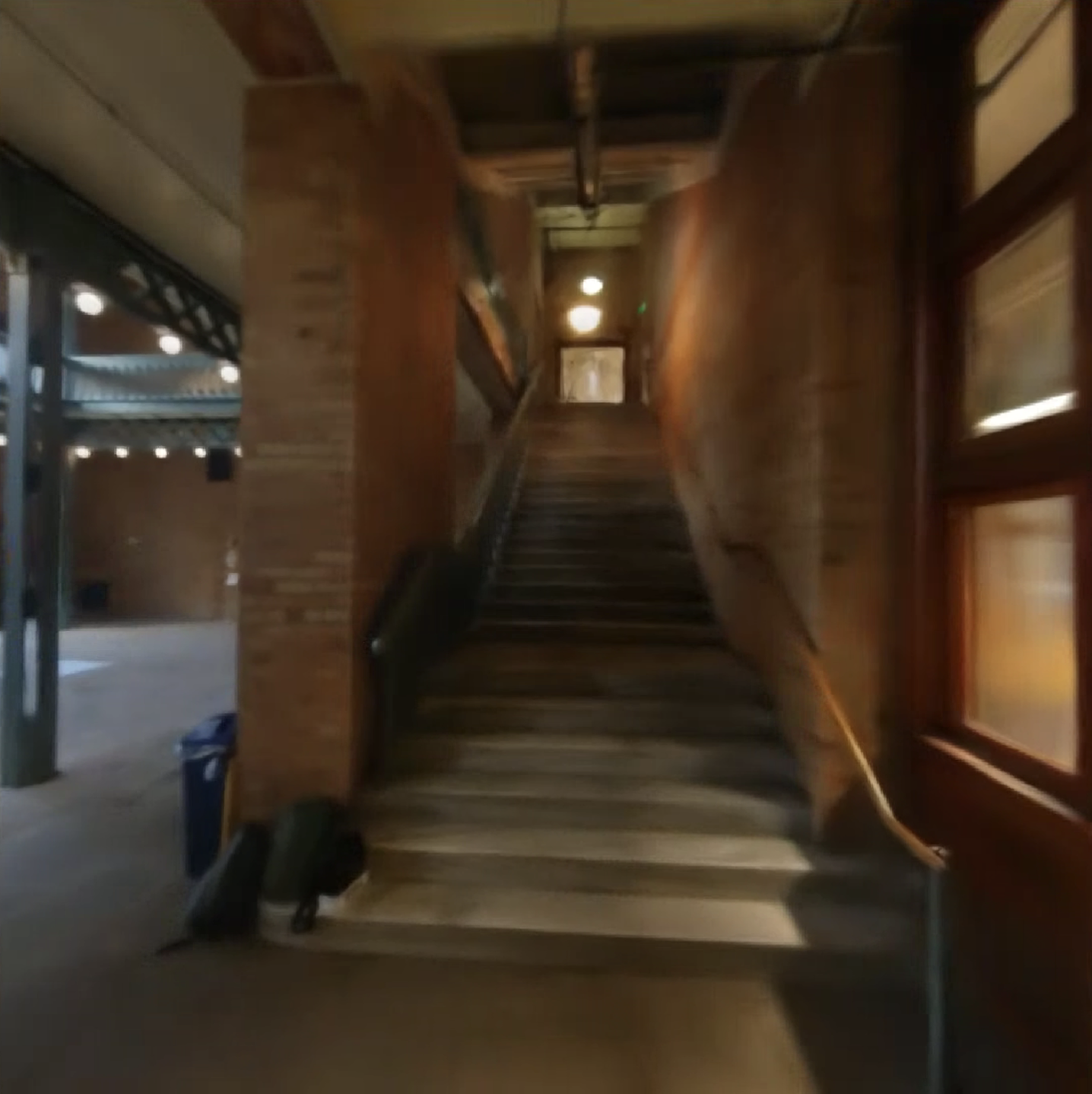}
  & \includegraphics[width=\hsize]{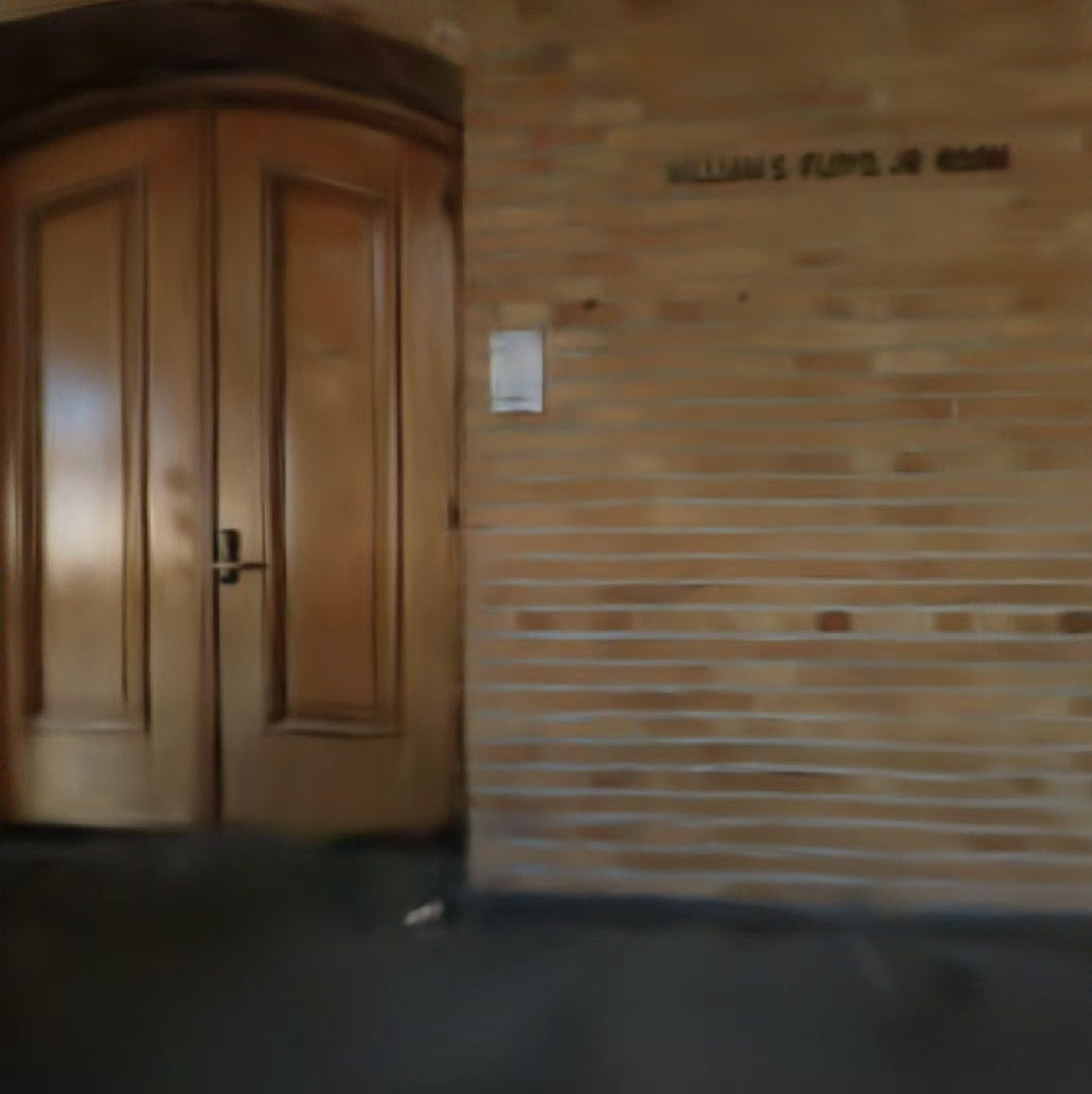}\\
\begin{subfigure}{0.05\linewidth} \end{subfigure} 
  & \includegraphics[width=\hsize]{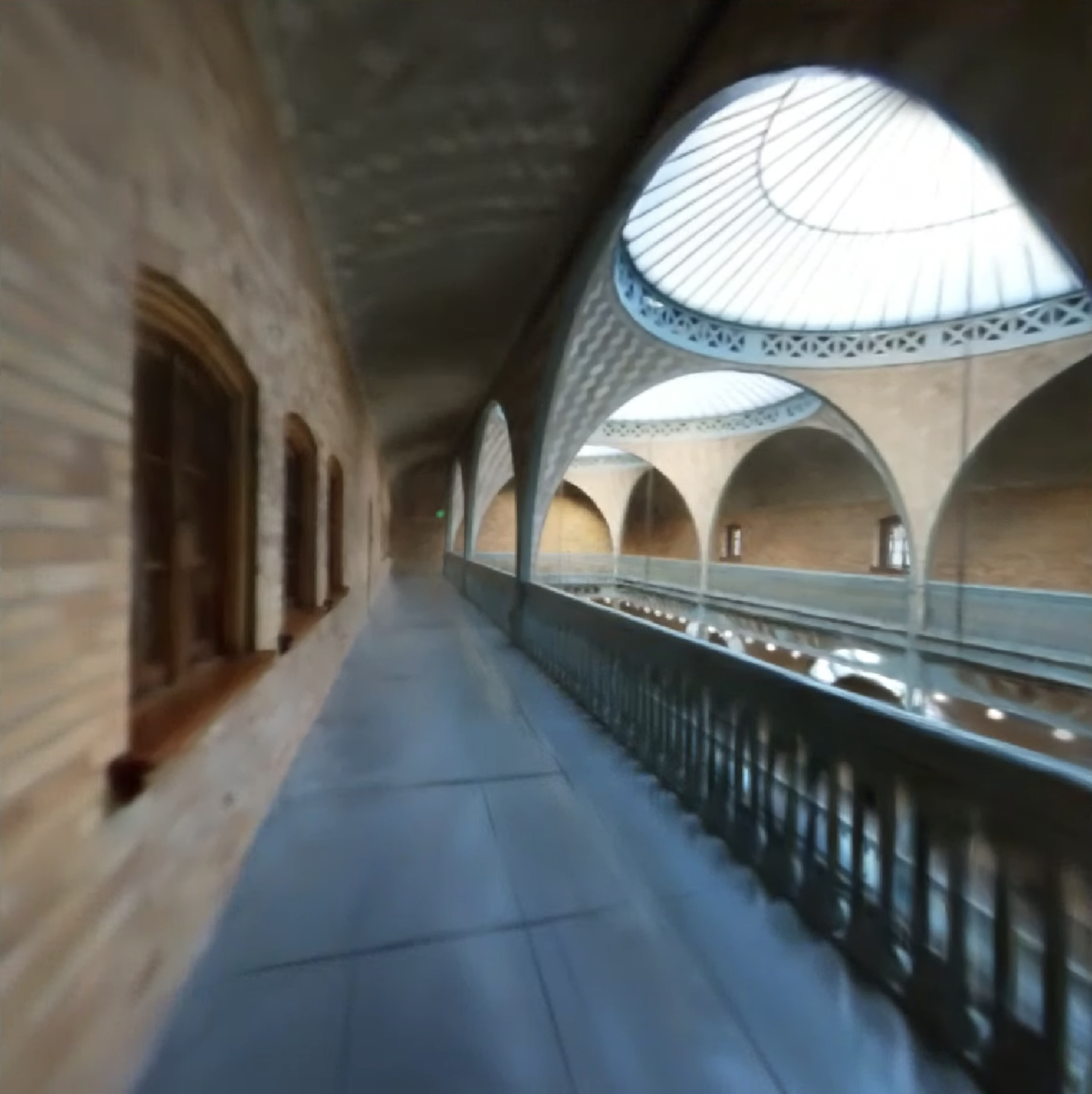}    
  & \includegraphics[width=\hsize]{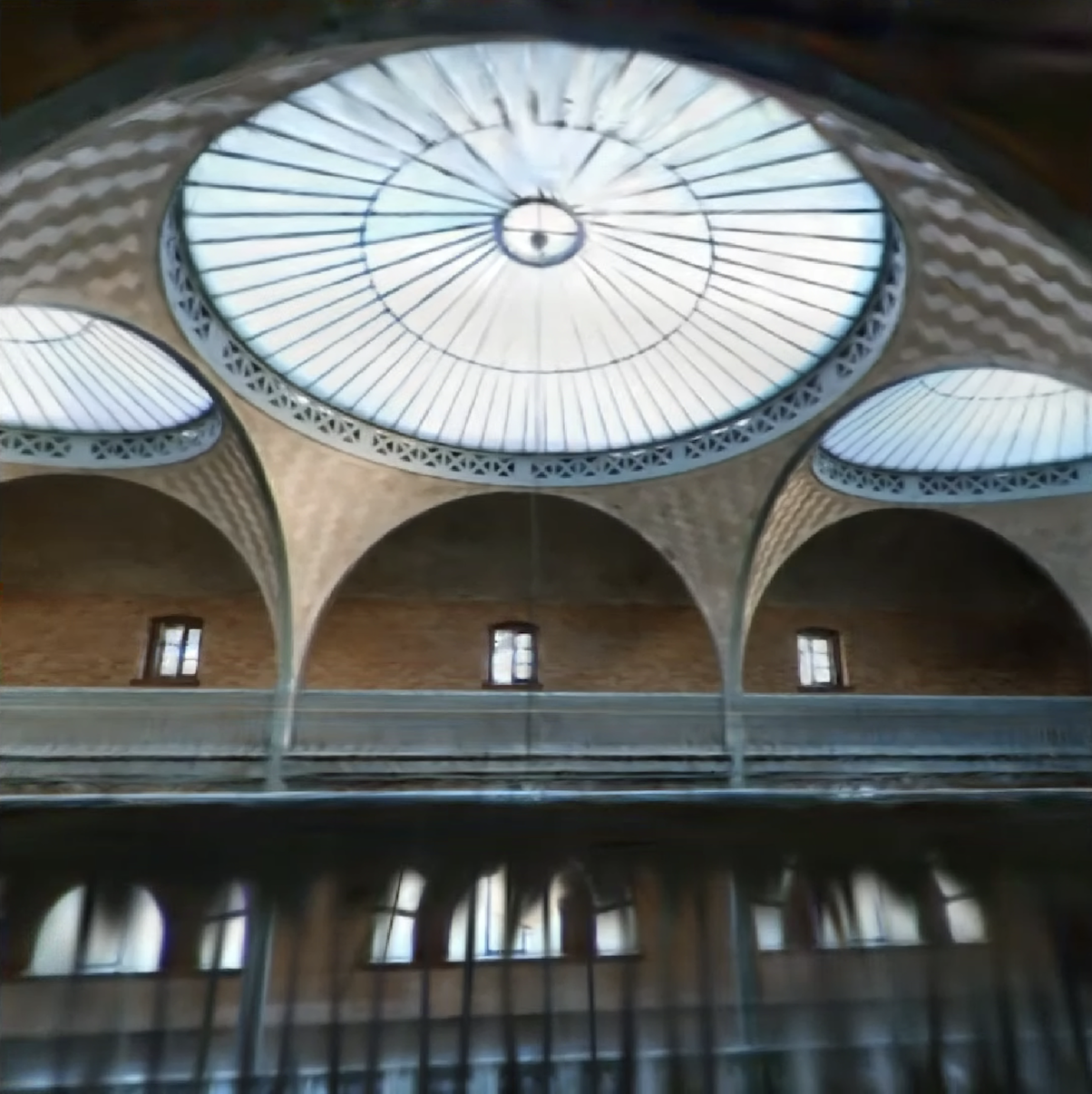}
  & \includegraphics[width=\hsize]{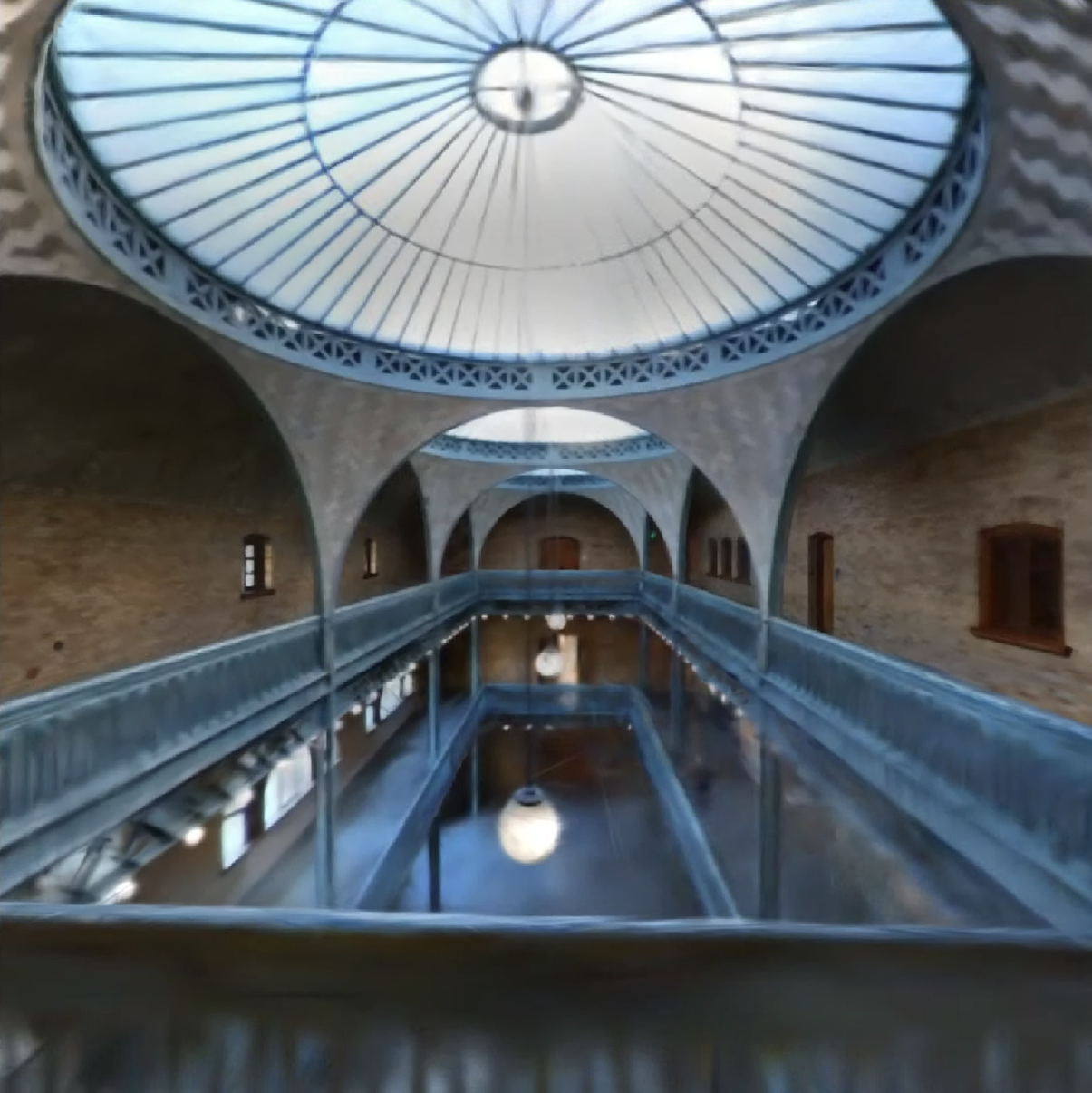}
  & \includegraphics[width=\hsize]{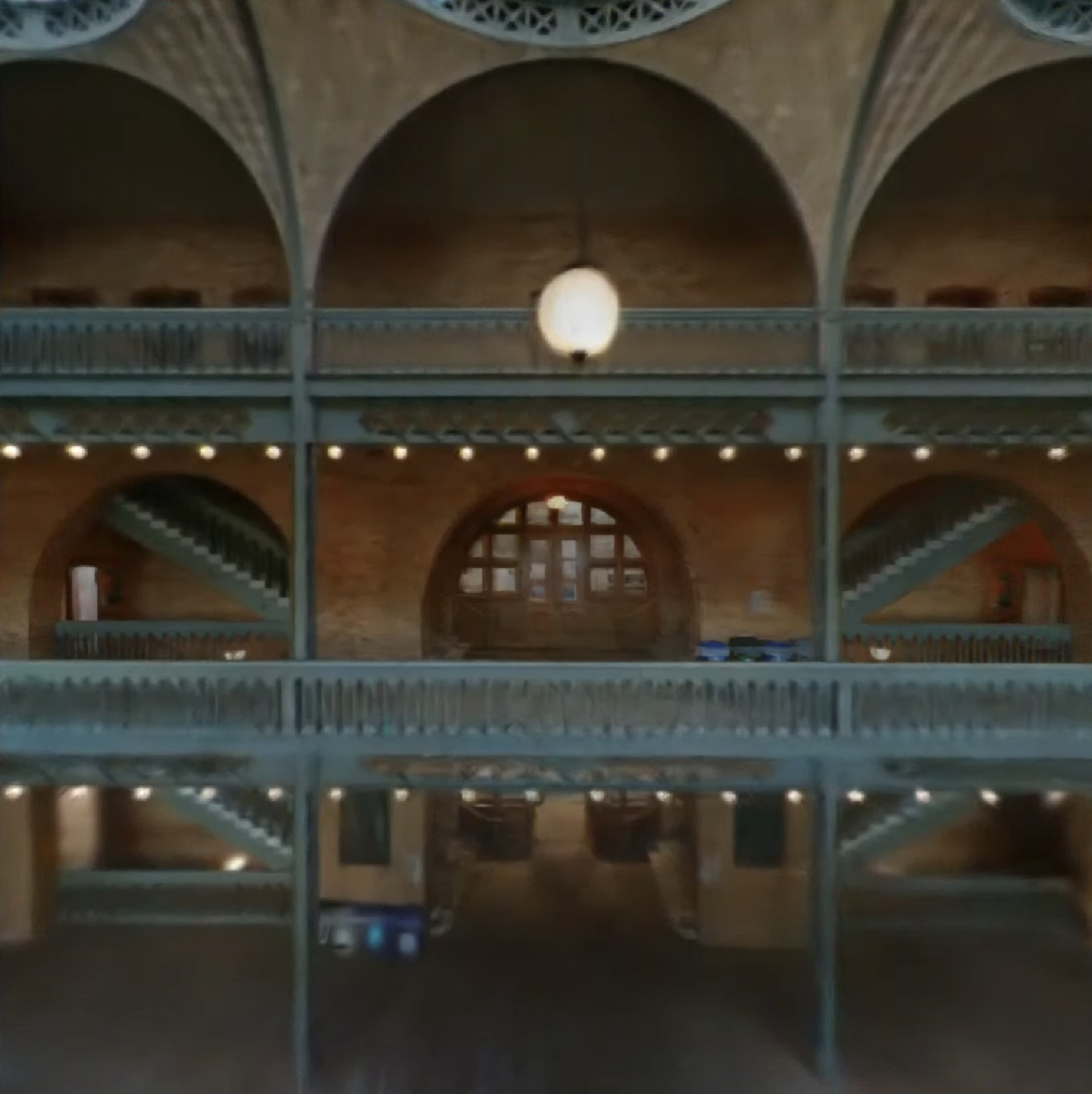}\\
\end{tabular}
\caption{Novel-view renderings of the Hearst Memorial Mining Building scene.}
\label{fig:hmmb_qualitative_results}
\end{figure}

\begin{figure}[htb]
\setlength\tabcolsep{1pt} 
\centering
\begin{tabular}{@{} r M{0.3\linewidth} M{0.3\linewidth} M{0.3\linewidth} @{}}
& (i) & (ii) & (iii) \\ \addlinespace
\begin{subfigure}{0.05\linewidth} \caption*{$T-1$}\label{} \end{subfigure} 
  & \includegraphics[width=\hsize]{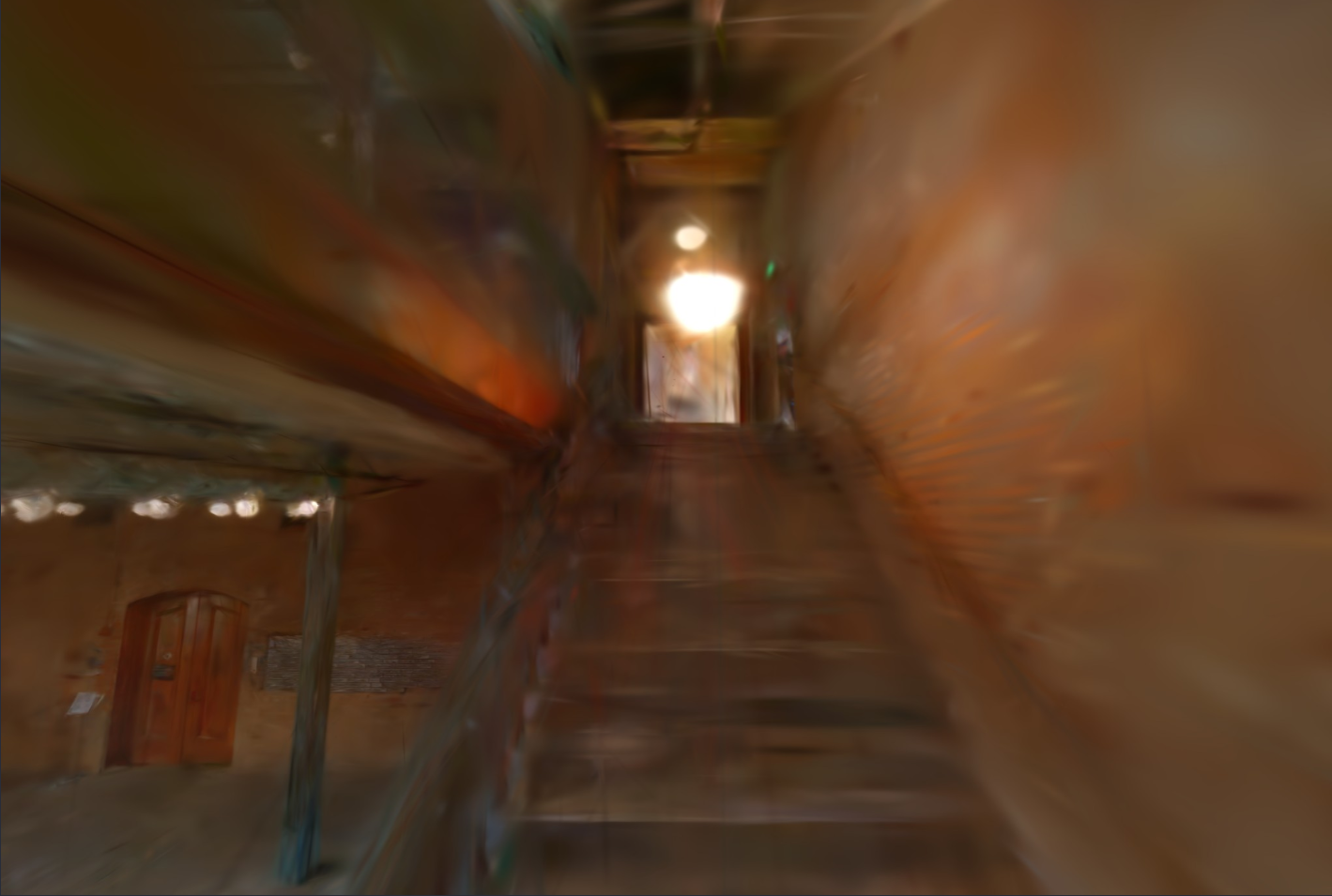} 
  & \includegraphics[width=\hsize]{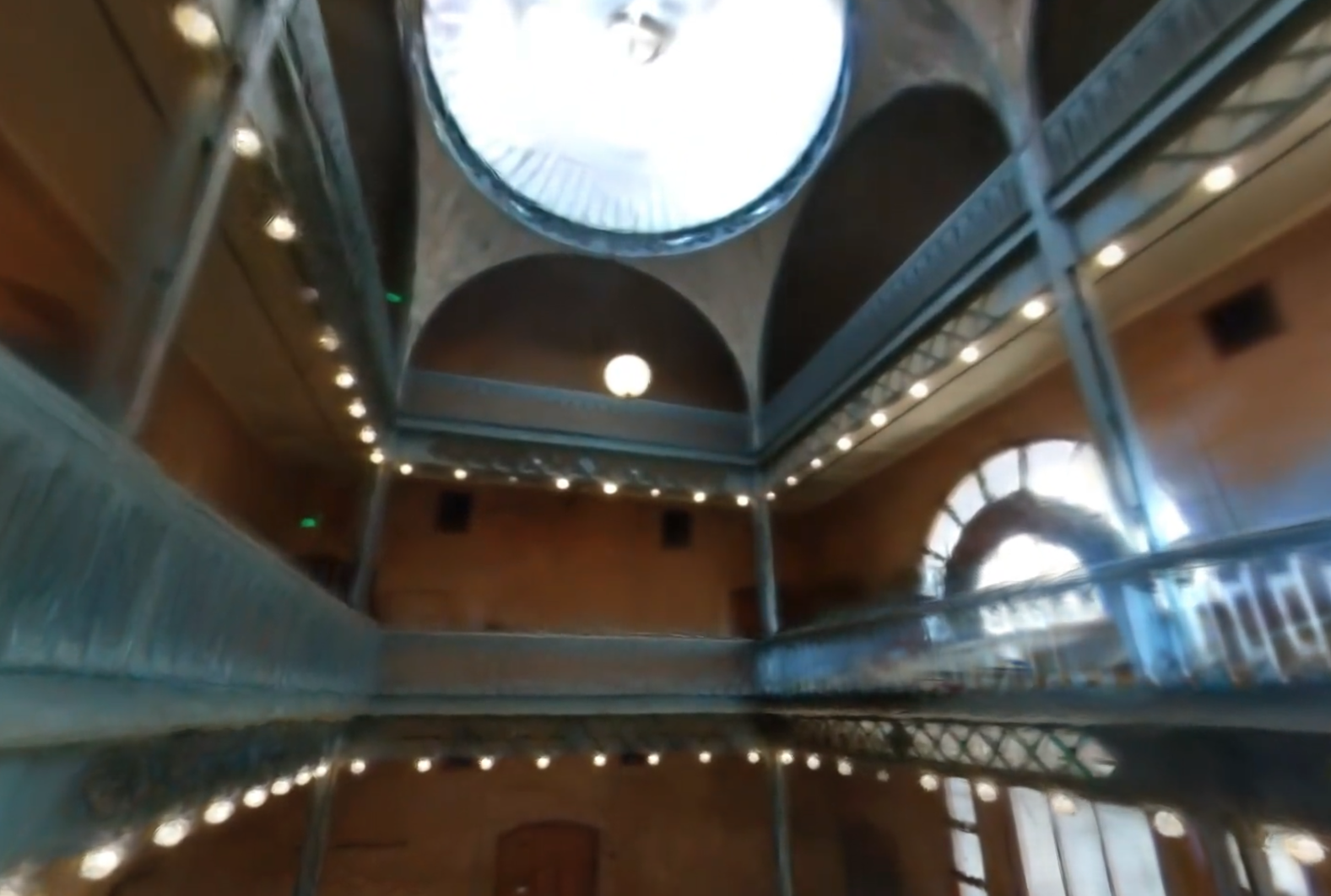}
  & \includegraphics[width=\hsize]{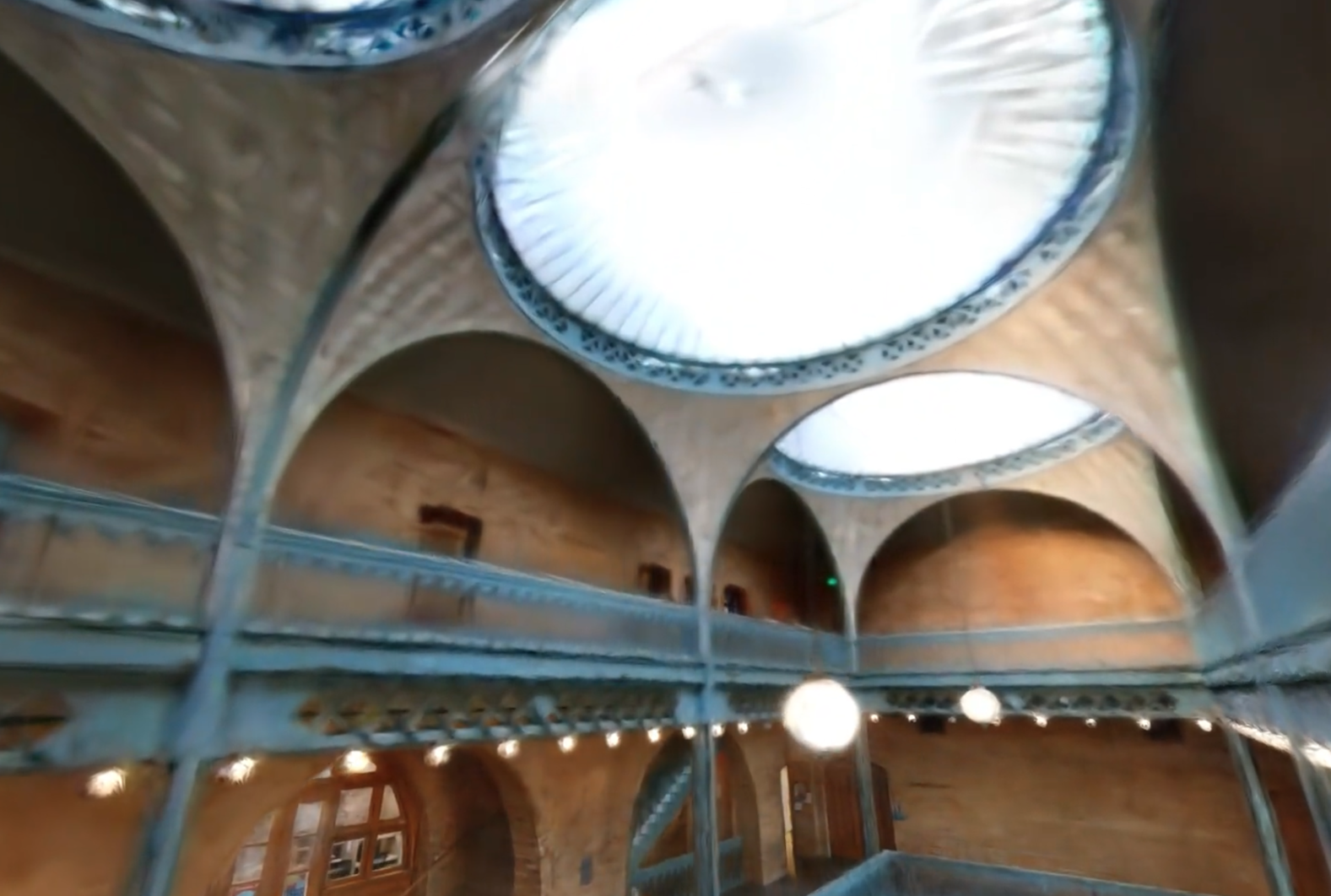}\\
\begin{subfigure}{0.05\linewidth} \caption*{$T$}\label{} \end{subfigure} 
  & \includegraphics[width=\hsize]{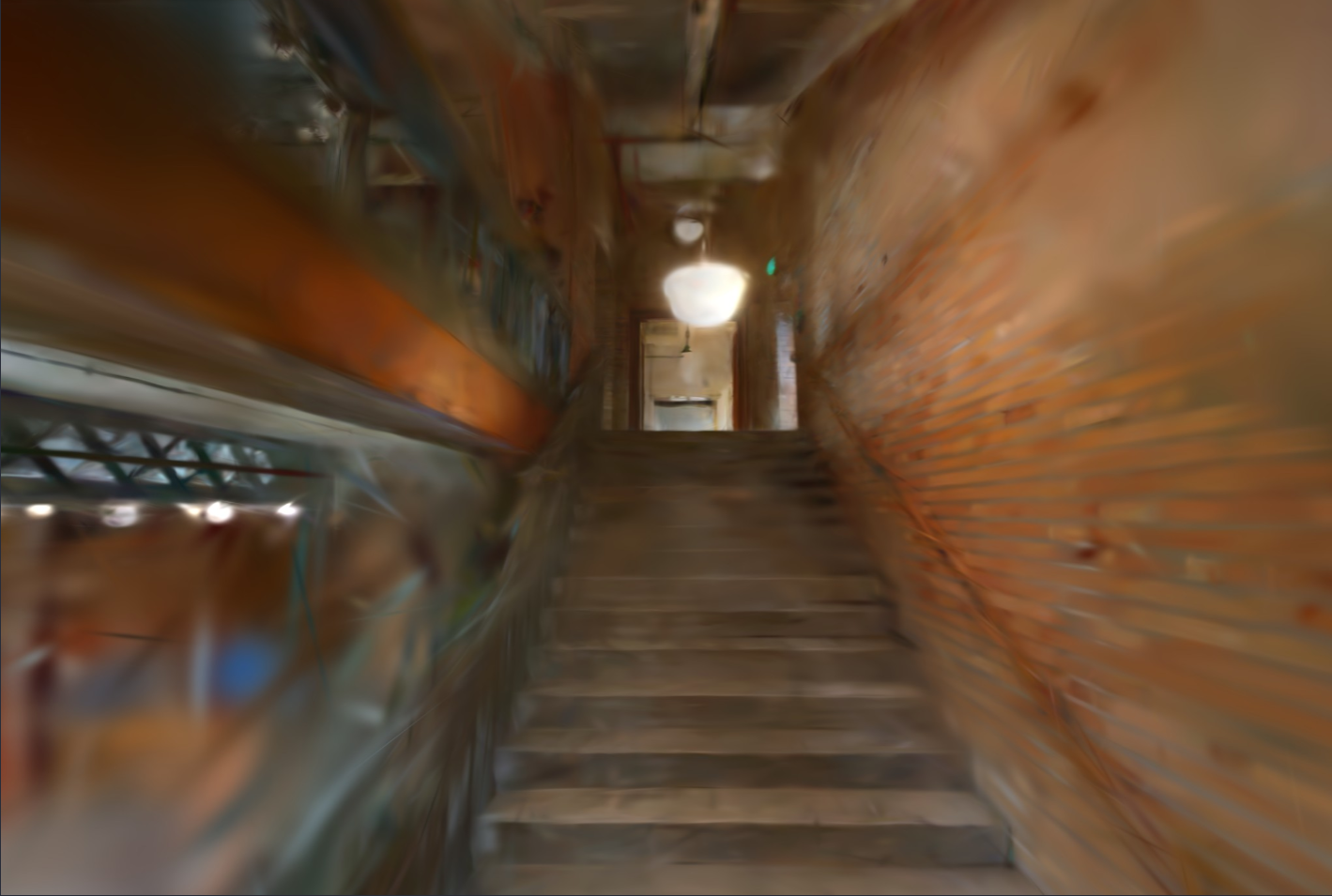}    
  & \includegraphics[width=\hsize]{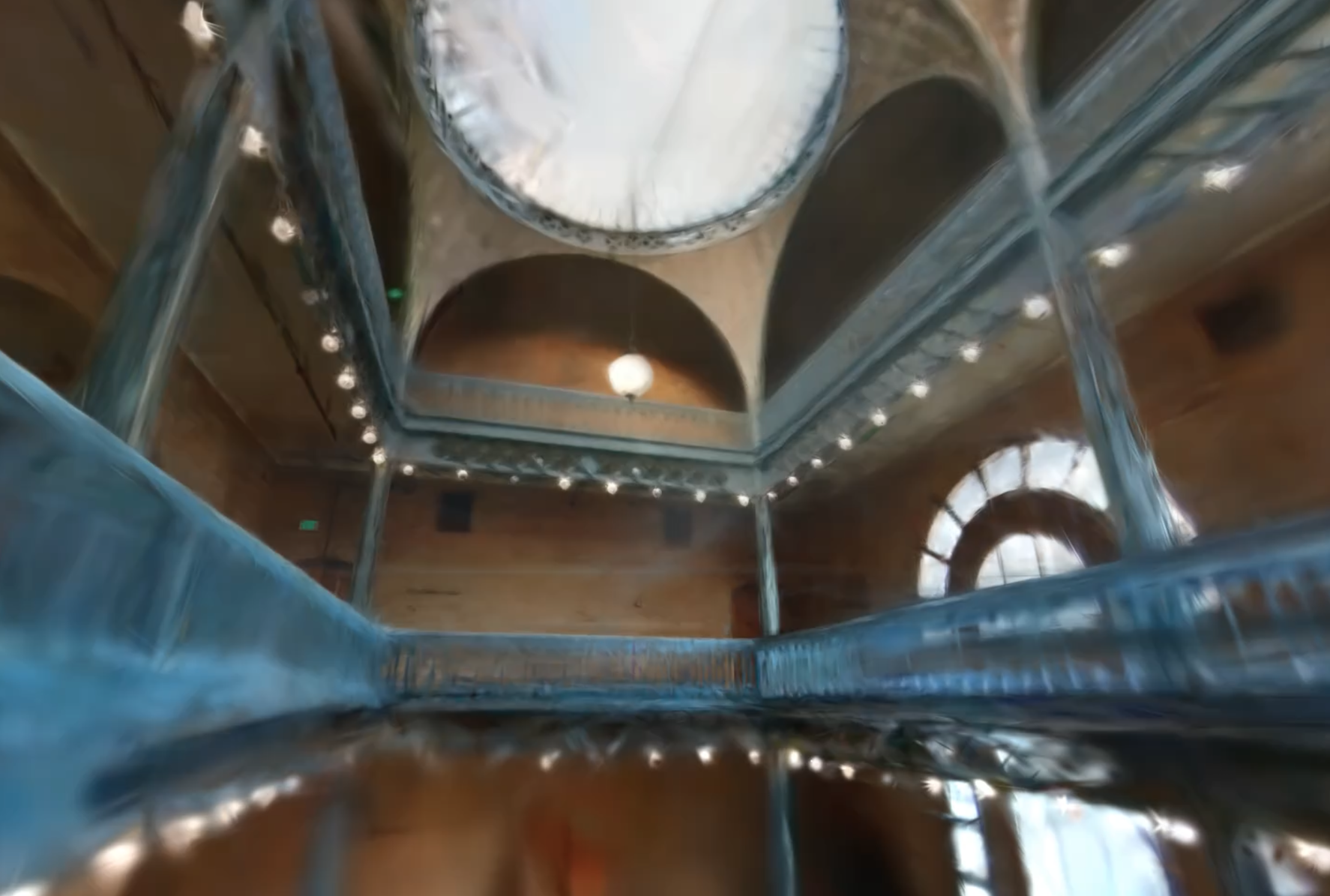}
  & \includegraphics[width=\hsize]{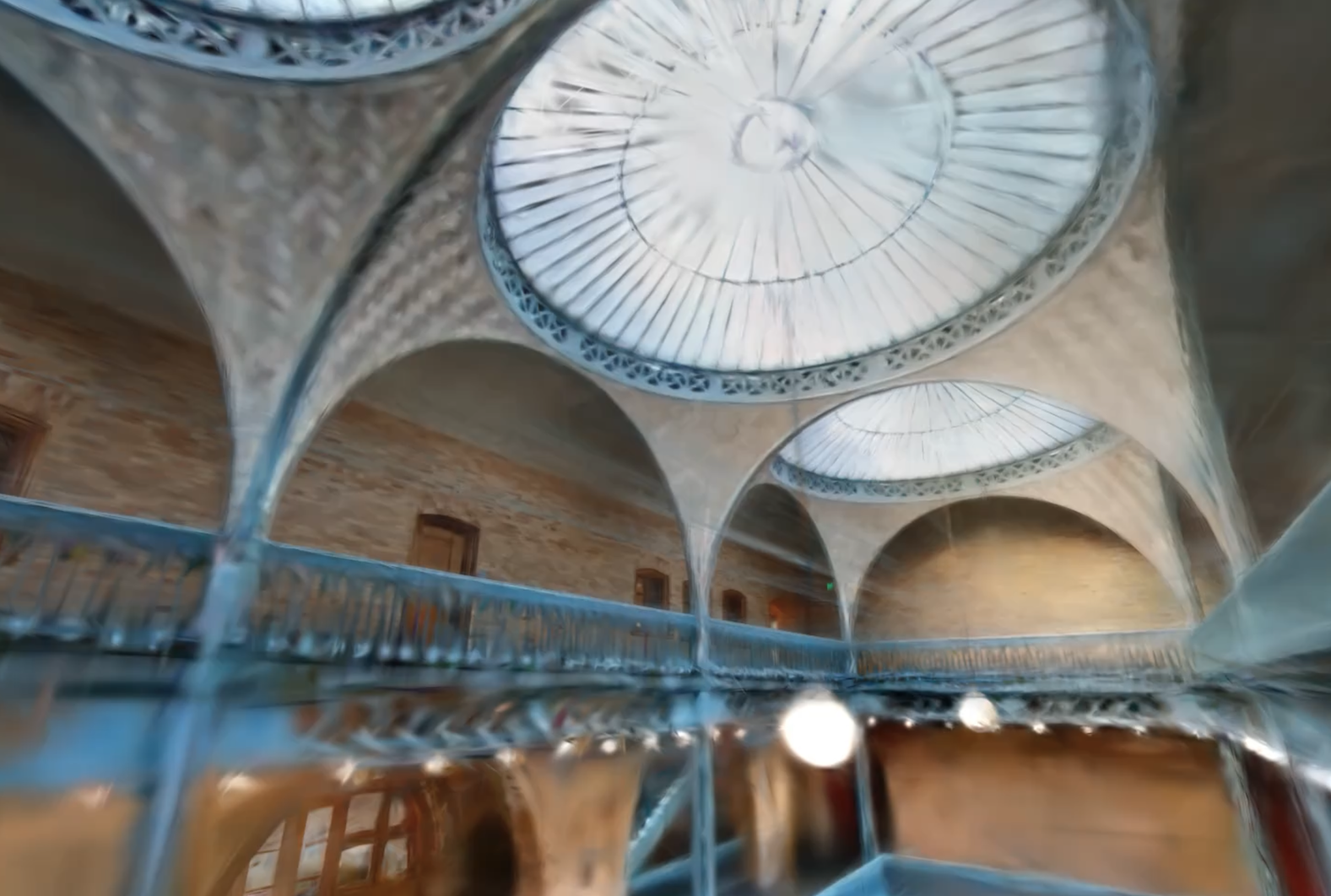}\\
\end{tabular}
\caption{Consecutive renderings at block boundaries of HMMB. \textbf{T - 1} corresponds to the previous floor and \textbf{T} corresponds to the next floor.}
\label{fig:hmmb_block_transition}
\end{figure}

We similarly compute PSNR and SSIM on the HMMB dataset, shown in \cref{tab:hmmb}. Here, time per-block shown in brackets can be parallelized due to our DAC method. The metrics for HMMB are lower than that of Cory Hall since HMMB is more complex compared to Cory Hall, with multiple floors, corridors and staircases. This increases the complexity of SfM, giving 3DGS a noisier initialization.
\cref{fig:hmmb_qualitative_results} shows our qualitative rendering of HMMB at different perspectives. \cref{fig:hmmb_block_transition} shows consecutive renderings at block boundaries of HMMB. As seen, our proposed coarse-to-fine alignment and on-demand block rendering technique for complex data captures successfully aligns neighboring blocks and loads the block almost instantly. Video rendering on the HMMB scene can be found at \cref{note:hmmb}.

\subsection{Ablation Studies}

\subsubsection{Inpainting}

\begin{table}[ht]
    \begin{center}
    {\resizebox{0.98\linewidth}{!}{
    \begin{tabular}{|c|c|c|c|c|c|c|c|c|c|c|}
         \hline
         \multirow{2}{*}{Method} & \multicolumn{2}{c|}{Block 0} & \multicolumn{2}{c|}{Block 1} & \multicolumn{2}{c|}{Block 2} & \multicolumn{2}{c|}{Block 3} & \multicolumn{2}{c|}{\textbf{Avg.}}\\
         \cline{2-11}
         & PSNR & SSIM & PSNR & SSIM & PSNR & SSIM & PSNR & SSIM & \textbf{PSNR} & \textbf{SSIM} \\
         \hline\hline
         Inpainting & 35.13 & 0.9651 & 33.76 & 0.9602 & 33.78 & 0.9645 & 35.89 & 0.9679 & \textbf{34.64} & \textbf{0.9644}\\
         \hline
         Masking & 34.10 & 0.9632 & 30.23 & 0.9521 & 30.24 & 0.9558 & 32.85 & 0.9620 & \textbf{31.85} & \textbf{0.9583}\\
         \hline
    \end{tabular}
    }}
    \end{center}
    \caption{Comparison of image quality between inpainting and masking on \textit{4S4B}.}
    \label{tab:inpaint_vs_mask}
\end{table}

We further compare our drone body inpainting approach with a simple drone masking approach, taking \textit{4S4B} as the target of comparison. For drone masking, we take the dilated drone body masks generated for inpainting, as described in \cref{sec:inpainting}, and input them into SfM to mask the drone body out when doing feature matching. Then, we modify the original 3DGS code, adding masking option for training to ignore the region with the drone body when computing loss. Quantitative results are presented in \cref{tab:inpaint_vs_mask}. As seen, on average, drone body inpainting achieves $2.78 dB$ higher PSNR and $0.0061$ higher SSIM than simply masking out the drone body. We reason that inpainting explicitly introduces pixel information across frames for novel-view optimization, while masking does not propagate information in the masked area.

\subsubsection{Block alignment error}

We evaluate the block alignment error of our coarse-to-fine method on \textit{4S4B} Scene 1 Cory Hall using root mean squared error (RMSE) in meters.
We compute their RMSE with respect to a ``ground truth'' obtained as follows: we input the frames of the overlapping regions from both blocks into COLMAP and hand-tune the COLMAP parameters to ensure a good reconstruction and pose recovery. RMSE is then computed per frame by taking the difference in the translation between the ground truth pose and the aligned pose.

\begin{table}[htb]
\begin{center}
\begin{tabular}{|c|c|c|c|c|c|c|}
    \hline
    \multirow{2}{*}{Method} & \multicolumn{4}{c|}{RMSE \textdownarrow (coarse + fine alignment)} & \multirow{2}{*}{Avg. Ratio\textdownarrow} & \multirow{2}{*}{Time\textdownarrow}\\
    \cline{2-5}
    & 0\textrightarrow 1 & 1\textrightarrow 2 & 2\textrightarrow 3 & Avg. & & \\
    \hline\hline
    CA-DC & 0.0399 m & 0.0577 m & 0.1442 m & 0.0806 m & 0.4344 & 20 min \\
    CA-FM & 0.0866 m & 0.0387 m & 0.1477 m & 0.0910 m & 0.4813 & 5 min \\
    \hline
\end{tabular}
\end{center}
\caption{Block alignment error on \textit{4S4B}. CA-FM: coarse alignment via feature matching, used in our pipeline. CA-DC: coarse alignment via direct COLMAP. RMSE is computed after coarse alignment and fine alignment for both methods.}
\label{tab:block_alignment_error}
\end{table}

Given our two-stage coarse-to-fine block alignment strategy, we experimented with two different methods in the coarse alignment stage shown in~\cref{tab:block_alignment_error}: (1) our proposed coarse alignment via feature matching as described in~\cref{sec:c2f_align}, called CA-FM, and (2) a direct coarse alignment strategy via COLMAP, which we call CA-DC.
CA-DC takes the last frame of the preceding block $f^{\text{prev}}_{\text{last}}$ and the frames in the overlapping region of the succeeding block $f^{\text{next}}_{i}$ and runs COLMAP to generate the relative pose of $f^{\text{prev}}_{\text{last}}$ to $f^{\text{next}}_{i}$. This relative pose provides a good coarse estimate of the transformation between the two blocks. We similarly compute the relative pose the other way round with $f^{\text{next}}_{\text{first}}$ and $f^{\text{prev}}_{i}$ and take the average between the two computed poses as the final coarse transformation. After coarse alignment, we run ICP as proposed in~\cref{sec:c2f_align}.

Our experiments show that while CA-DC has a lower RMSE score than CA-FM, it is much slower than CA-FM since it performs SfM to estimate relative poses.
We also include an average ratio metric to the aid understanding of the RMSE, which we define as the ratio of the alignment error to the average distance between each successive image frame. For both methods, the average ratio is less than $0.5$, which means that the aligned frames are less than half a frame off their ground truth pose.
Since the difference in RMSE score and average ratio between the two methods are minimal, we use CA-FM as our coarse alignment strategy.

\subsection{Limitations}

One limitation is that SfM requires careful parameter adjustment and our pipeline time and novel-view synthesis result is partly bottle-necked by its performance. 
Another limitation comes from inpainting areas with high image frequencies. The inpainted result might not be multi-view accurate, producing visual artifacts in the reconstructed Gaussian Splat.
Moreover, at block boundaries, color differences may occur since each block is trained separately, and viewing discrepancies can arise if occlusion is present, as in \cref{fig:hmmb_block_transition}. Image blending techniques with appropriate weights \cite{tancik2022blocknerf, chen2023scalarnerf} could be applied to improve render quality at block boundaries. Lastly, at indoor corners where wall textures are plain and poorly lit, there could still be erroneous feature matching, leading to foggy rendering. Post-processing methods could be investigated to mitigate this problem.

\section{Conclusions} 
\label{sec:conclusion}

We present an efficient and scalable pipeline for large-scale indoor novel-view synthesis, starting from data capture to scene reconstruction. Our usage of the 360\textdegree \,camera captures diverse viewpoints, making drone piloting and data capture much easier than with a standard perspective camera. By utilizing a divide-and-conquer strategy, our pipeline is parallelizeable and scales up to large complex indoor scenes. Our experiments show marked improvement in both quality and speed over prior methods on the same scene. In addition, we evaluated our pipeline on a large, multi-story, building-scale scene and achieved photorealistic reconstruction results, demonstrating our pipeline's overall effectiveness.

%
%
\bibliographystyle{splncs04}
\bibliography{main}
\end{document}